\PassOptionsToPackage{table}{xcolor}
\documentclass{article} 
\usepackage{iclr2024_conference,times}

\usepackage{amsmath,amsfonts,bm}

\def\eqref#1{equation~\ref{#1}}

\def\1{\bm{1}}

\DeclareMathAlphabet{\mathsfit}{\encodingdefault}{\sfdefault}{m}{sl}
\SetMathAlphabet{\mathsfit}{bold}{\encodingdefault}{\sfdefault}{bx}{n}

\newcommand{\ie}{\textit{i.e.}}
\newcommand{\eg}{\textit{e.g.}}

\usepackage{graphicx}
\usepackage{subcaption}
\usepackage{amssymb}
\usepackage{enumitem}
\usepackage{wrapfig}
\usepackage{amsmath}
\usepackage{bbm}

\usepackage{booktabs}
\usepackage{xcolor}
\usepackage{bm}
\usepackage{multirow}
\usepackage[ruled,vlined,noend]{algorithm2e}
\PassOptionsToPackage{table}{xcolor}
\DeclareMathAlphabet\mathbfcal{OMS}{cmsy}{b}{n}

\definecolor{mygray}{gray}{0.9}

\usepackage{hyperref}
\usepackage{url}

\title{Exploring Diffusion Time-steps for Unsupervised Representation Learning}

\author{%
 \textbf{Zhongqi Yue}\textsuperscript{1}, \textbf{Jiankun Wang}\textsuperscript{1},  \textbf{Qianru Sun}\textsuperscript{2},  \textbf{Lei Ji}\textsuperscript{3},  \textbf{Eric I-Chao Chang}\textsuperscript{3},  \textbf{Hanwang Zhang}\textsuperscript{1}\\
{\small \textsuperscript{1}Nanyang Technological University,\quad \textsuperscript{2}Singapore Management University,\quad \textsuperscript{3}Microsoft Research Asia}\\
{\tt\small yuez0003@ntu.edu.sg,\quad jiankun001@e.ntu.edu.sg,\quad qianrusun@smu.edu.sg},\\
\tt\small leiji@microsoft.com,\quad eric.i.chang@outlook.com,\quad hanwangzhang@ntu.edu.sg}

\iclrfinalcopy 
\begin{document}

\maketitle

\begin{abstract}

Representation learning is all about discovering the hidden modular attributes that generate the data faithfully. We explore the potential of Denoising Diffusion Probabilistic Model (DM) in unsupervised learning of the modular attributes. We build a theoretical framework that connects the diffusion time-steps and the hidden attributes, which serves as an effective inductive bias for unsupervised learning. Specifically, the forward diffusion process incrementally adds Gaussian noise to samples at each time-step, which essentially collapses different samples into similar ones by losing attributes, \eg, fine-grained attributes such as texture are lost with less noise added (\ie, early time-steps), while coarse-grained ones such as shape are lost by adding more noise (\ie, late time-steps). To disentangle the modular attributes, at each time-step $t$, we learn a $t$-specific feature to compensate for the newly lost attribute, and the set of all $\{1,\ldots,t\}$-specific features, corresponding to the cumulative set of lost attributes, are trained to make up for the reconstruction error of a pre-trained DM at time-step $t$. On CelebA, FFHQ, and Bedroom datasets, the learned feature significantly improves attribute classification and enables faithful counterfactual generation, \eg, interpolating only one specified attribute between two images, validating the disentanglement quality. Codes are in \url{https://github.com/yue-zhongqi/diti}.

\end{abstract}
\section{Introduction}
\label{sec:1}

A good feature representation should faithfully capture the underlying generative attributes in a compact and modular vector space~\cite{bengio2013representation}, enabling not only sample inference (\eg, image classification) but also counterfactual generation~\cite{besserve2018counterfactuals} (\eg, image synthesis of unseen attribute combinations).
Over the past decade, discriminative training has been the feature learning mainstream with exceptional performance in inference tasks~\cite{he2016deep,he2020momentum}. However, it hardly achieves faithful generation due to the deliberate discard of certain attributes, \eg, class-irrelevant ones in supervised learning or augmentation-related ones in contrastive learning.

On the other hand,  generative Denoising Diffusion Probabilistic Model (DM)~\cite{sohl2015deep,song2020score} can retain all the underlying attributes for faithful generation~\cite{dhariwal2021diffusion}, or even extrapolate novel attribute combinations by textual control~\cite{rombach2022high} (\eg, ``Teddy bear skating in Time Square''),
outperforming other generative models like VAE~\cite{kingma2014ICLR} and  GAN~\cite{creswell2018generative}. This implies that DM effectively captures the modularity of hidden attributes~\cite{yue2021counterfactual}.
However, as DM's formulation has no explicit encoders that transform samples into feature vectors, the conventional encoder-decoder feature learning paradigm via reconstruction~\cite{higgins2017beta} is no longer applicable.

\begin{figure}[t!]
    \centering 
    \captionsetup{font=footnotesize,labelfont=footnotesize,skip=5pt}
    \begin{subfigure}[t]{1.0\linewidth}
         \includegraphics[width=1.0\linewidth]{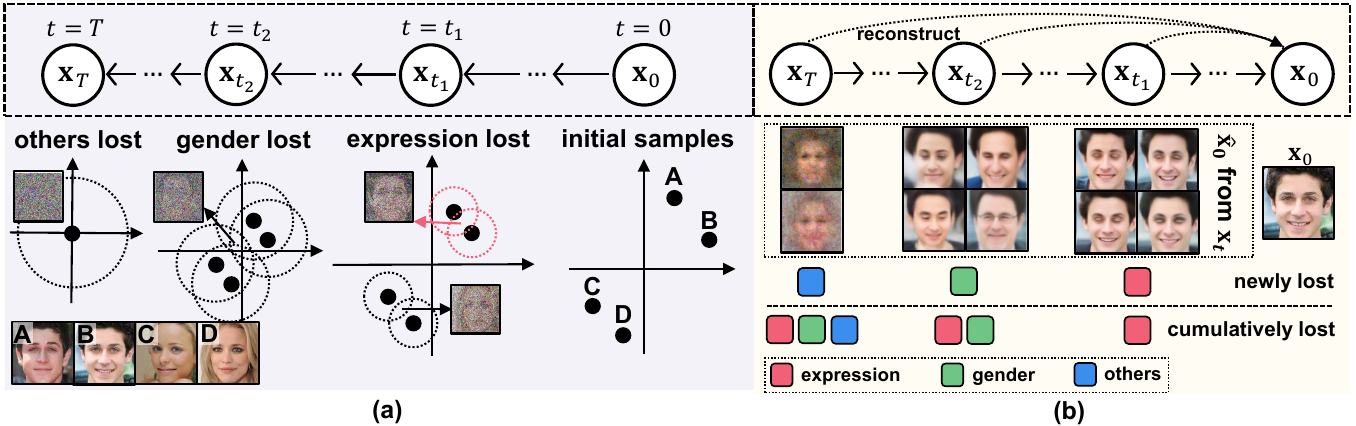}
         \phantomcaption
         \label{fig:1a}
    \end{subfigure}
    \begin{subfigure}[t]{0\linewidth} 
         \includegraphics[width=0\linewidth]{example-image-b}
         \phantomcaption
         \label{fig:1b}   
    \end{subfigure}
    \vspace*{-8mm}
    \caption{(a) Illustration of attribute loss as time-step $t$ increases in the forward diffusion process. The two axes depict a two-dimensional sample space. (b) DM reconstructed $\mathbf{x}_0$, denoted as $\hat{\mathbf{x}}_0$, from randomly sampled $\mathbf{x}_t$ at various $t$. DM is pre-trained on CelebA, from where $\mathbf{x}_0$ is drawn.}
    \label{fig:1}
    \vspace*{-4mm}
\end{figure}

Can we extract DM's knowledge about the modular attributes as a compact feature vector?
To answer the question, we start by revealing a natural connection between the diffusion time-steps and the hidden attributes. In the forward diffusion process, given an initial sample $\mathbf{x}_0$, a small amount of Gaussian noise $\mathcal{N}(\mathbf{0},\mathbf{I})$ is incrementally added to the sample across $T$ time-steps, resulting in a sequence of noisy samples $\mathbf{x}_1,\ldots,\mathbf{x}_T$. In particular, each $\mathbf{x}_t$ adheres to its noisy sample distribution $q(\mathbf{x}_t|\mathbf{x}_0)$, whose mean and covariance matrix $\sigma^2 \mathbf{I}$ is denoted as a dot and a circle with radius $\sigma$ in Figure~\ref{fig:1a}, respectively.
As $t$ increases, $q(\mathbf{x}_t|\mathbf{x}_0)$ progressively collapses towards a pure noise by reducing mean (\ie, dots moving closer to the origin), and increasing $\sigma$ (\ie, larger radii), which enlarges the overlapping area (OVL) of the probability density function of noisy sample distributions.
In particular, we theoretically show in Section~\ref{sec:4.1} that OVL is correlated with DM's ambiguous reconstruction of different samples, \eg, due to the large overlap between the red circles at $t_1$, noisy samples drawn from the two distributions are reconstructed ambiguously as either A or B, losing the attribute ``expression''\footnote{Attribute name is illustrative. Our method uses no attribute supervision, nor explicitly names them.} that distinguishes them.
As $t$ increases, more noisy sample distributions exhibit significant overlaps, inducing additional attribute loss, \eg, further losing ``gender'' at $t_2$ that confuses the identities of A/B and C/D.
Eventually, with a large enough $T$, $\mathbf{x}_T$ becomes pure noise and all attributes are lost---the reconstruction can be any sample.

This intrinsic connection between diffusion time-steps and modular attributes can act as an effective inductive bias for unsupervised learning.
Specifically, DM training can be viewed as learning to reconstruct $\mathbf{x}_0$ given $\mathbf{x}_t$ at each $t$ (Section~\ref{sec:3.2}).
However, due to the aforementioned attribute loss, perfect DM reconstruction is impossible, \eg, in Figure~\ref{fig:1b}, reconstructions exhibit variations in the cumulatively lost attributes.
Hence, by contraposition, if a feature enables perfect reconstruction, it must capture the cumulatively lost attributes.
This motivates us to learn an \emph{expanding} set of features to supplement the \emph{expanding} lost attributes as $t$ increases.
Specifically, we first map $\mathbf{x}_0$ to a feature $\mathbf{z}=f(\mathbf{x}_0)$ using an encoder $f$. Then we partition $\mathbf{z}$ into $T$ disjoint subsets $\{\mathbf{z}_{i}\}_{i=1}^T$.
At each $t$, we optimize $f$ so that $\{\mathbf{z}_{i}\}_{i=1}^t$ compensates for a pre-trained DM's reconstruction error.
Intuitively by induction, starting from $t=0$ with no lost attribute; if $\{\mathbf{z}_{i}\}_{i=1}^t$ captures the cumulatively lost attributes till $t$, $\mathbf{z}_{t+1}$ must capture the newly lost attribute to enable perfect reconstruction at $t+1$; until $t=T-1$, $\{\mathbf{z}_{i}\}_{i=1}^{T-1}\cup \mathbf{z}_{T}$ learns all hidden attributes in a compact and modular vector space.

We term our approach as \textbf{DiTi} to highlight our leverage of \underline{Di}ffusion \underline{Ti}me-step as an inductive bias for feature learning. We summarize the paper structure and our contributions below:
\begin{itemize}[leftmargin=+0.25in,itemsep=3pt,topsep=0pt,parsep=0pt]
    \item In Section~\ref{sec:3.1}, we formalize the notion of good feature with a definition of disentangled representation~\cite{higgins2018towards} and provide a preliminary introduction to DM in Section~\ref{sec:3.2}.
    \item In Section~\ref{sec:4.1}, we build a theoretical framework that connects diffusion time-steps and the hidden modular attributes, which motivates a simple and practical approach for unsupervised disentangled representation learning, discussed in Section~\ref{sec:4.2}.
    \item In Section~\ref{sec:5}, by extensive experiments on CelebA~\cite{liu2015faceattributes}, FFHQ~\cite{karras2019style} and Bedroom~\cite{yu2015lsun}, our DiTi feature brings significant improvements in attribute inference accuracy and enables counterfactual generation, verifying its disentanglement quality.
\end{itemize}
\section{Related Works}
\label{sec:2}

\noindent\textbf{DM for Representation Learning}. There are three main approaches: 1) DDIM~\cite{song2020denoising} inversion aims to find a latent $\mathbf{x}_T$ as feature, by DDIM sampling with the learned DM, can reconstruct the image $\mathbf{x}_0$. However, this process is time-consuming, and the resulting latent is difficult to interpret~\cite{hertz2022prompt}. 2) ~\cite{kwon2022diffusion} uses the bottleneck feature of the U-Net at all time-steps. However, the feature is not compact and difficult for downstream leverage. 3) The closest to our works are~\cite{preechakul2022diffusion, zhang2022unsupervised}. However, they learn a \emph{time-step-independent} $\mathbf{z}$ to supplement DM reconstruction error. Without explicit design to enforce modularity, the learned feature entangles all the attributes. At time-step $t$, it leaks information about attributes lost at a later time-step (\ie, $t+1,\ldots,T$). Hence, instead of learning the lost attribute $\mathbf{z}_t$, the reconstruction may digress to use the spurious correlation between leaked information and lost attribute. In Section~\ref{sec:5}, we validate this deficiency by observing failed counterfactual generation and lower attribute prediction accuracy.

\noindent\textbf{Disentangled Representation} separates the modular hidden attributes that generate data, which can be formally defined from a group-theoretical view~\cite{higgins2018towards}.
Our work focuses on the unsupervised learning setting.
Existing methods typically learn a VAE~\cite{burgess2018understanding} or GAN~\cite{jeon2021ib} with information bottleneck.
Yet they lack explicit inductive bias that is necessary for unsupervised disentanglement~\cite{locatello2019challenging}.
Our work reveals that the connection between hidden attributes and diffusion time-steps is an effective inductive bias towards disentanglement (Section~\ref{sec:4.1}).

\section{Preliminaries}
\label{sec:3}

\begin{figure}[t!]
    \centering
    \captionsetup{font=footnotesize,labelfont=footnotesize,skip=5pt}
    \includegraphics[width=1.0\linewidth]{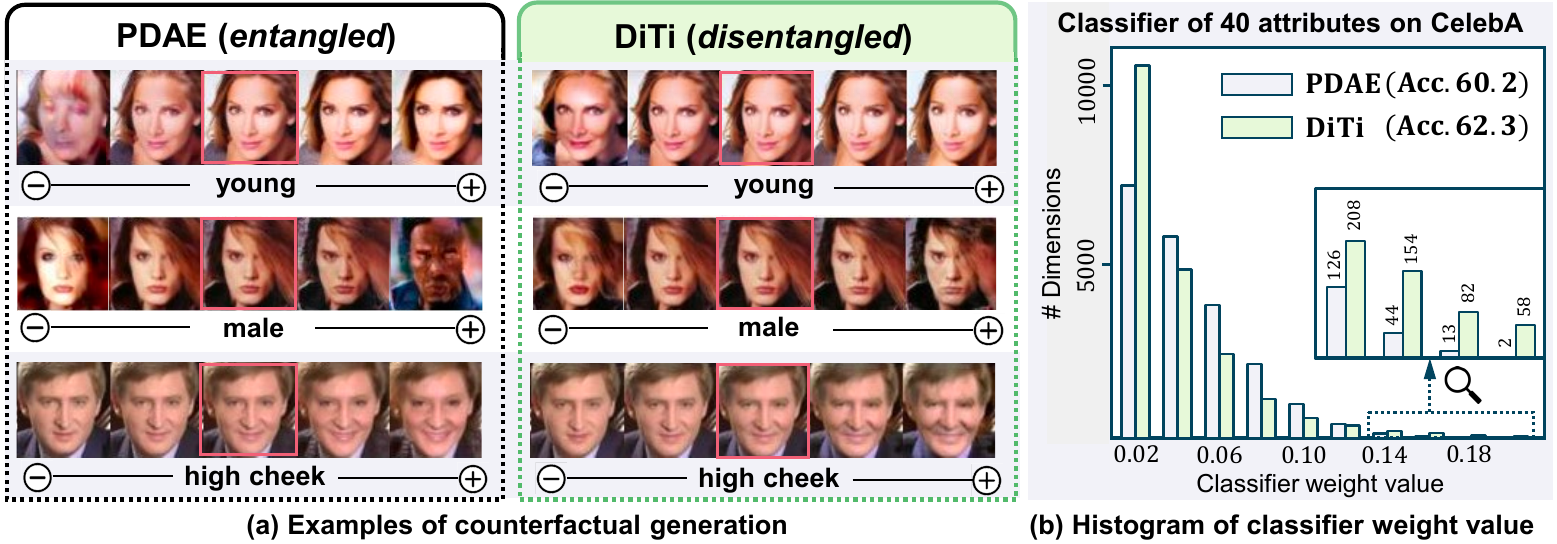}
    \caption{(a) Counterfactual generations on CelebA by manipulating 16 out of 512 feature dimensions (\ie, simulating the edit of a single $\mathbf{z}_i$). A disentangled representation enables editing a single attribute (\eg, gender) without affecting others (\eg, lighting) and promotes faithful extrapolation (\eg, no artifacts). (b) Histogram of the classifier weight value. More dimensions of DiTi weights are closed to $1$ and $0$ (explanations in the text).}
    \label{fig:manipulate}
    \vspace*{-5mm}
\end{figure}

\subsection{Disentangled Representation}
\label{sec:3.1}

A good feature can be formally defined by the classic notion of disentangled representation. We aim to provide a preliminary introduction below, and refer interesting readers to~\cite{higgins2018towards, wang2021self} for more formal details.
Each sample $\mathbf{x}\in \mathcal{X}$ in the image space $\mathcal{X}$ is generated from a hidden attribute vector $\mathbf{z}\in \mathcal{Z}$ in the vector space $\mathcal{Z}$ through an injective mapping $\Phi:\mathcal{Z}\to \mathcal{X}$ (\ie, different samples are generated by different attributes).
In particular, $\mathcal{Z}$ can be decomposed into the Cartesian product of $N$ subspaces $\mathcal{Z}=\mathcal{Z}_1\times \ldots \times \mathcal{Z}_N$, where each $\mathcal{Z}_i$ is called a \emph{modular} attribute, as its value can be locally intervened without affecting other modular attributes (\eg, changing ``expression'' without affecting ``gender'' or ``age'').
A disentangled representation is a mapping $f:\mathcal{X}\to \mathcal{Z}$ that inverses $\Phi$, \ie, given each image $\mathbf{x}$, $f(\mathbf{x})$ recovers the modular attributes as a feature vector $\mathbf{z}=[\mathbf{z}_1,\ldots,\mathbf{z}_N]$, which possesses desired properties of a good feature in the following common views:

\noindent\textbf{Counterfactual generation}. A counterfactual image of $\mathbf{x}$ by intervening the $i$-th modular attribute with $g_i$ (\eg, \textit{aging}) is denoted as $g_i \cdot \mathbf{x} \in \mathcal{X}$ (\eg, an \textit{older} $\mathbf{x}$). Specifically, $g_i \cdot \mathbf{x}$ is generated by first mapping $\mathbf{x}$ to $\mathbf{z}$ by $f$, editing $\mathbf{z}_i$ according to $g_i$ without changing other attributes, and finally feeding the modified $\mathbf{z}$ to $\Phi$.
In fact, $g_i \cdot \mathbf{x}$ is formally known as group action~\cite{Judson1994Abstract}. Figure~\ref{fig:manipulate} shows counterfactual images generated by editing three attributes.

\noindent\textbf{Sample inference}. A modular feature facilitates robust inference in downstream tasks about \emph{any} subset of $\{\mathbf{z}_i\}_{i=1}^N$, \eg, a linear classifier can eliminate task-irrelevant environmental attributes by assigning larger values to the dimensions corresponding to $\mathbf{z}_i$ and $0$ to other dimensions. In Figure~\ref{fig:manipulate}, by explicitly disentangling the modular attributes (Section~\ref{sec:4}), our DiTi achieves higher accuracy in attribute prediction than PDAE~\cite{zhang2022unsupervised} (without disentangling), while more dimensions of our classifier weight are closed to $1$ (corresponding to $\mathbf{z}_i$) and $0$ (environmental attributes).

\subsection{Denoising Diffusion Probabilistic Model}
\label{sec:3.2}

Denoising Diffusion Probabilistic Model~\cite{ho2020denoising} (DM) is a latent variable model with a fixed \emph{forward process}, and a learnable \emph{reverse process}. Details are given below.

\noindent\textbf{Forward Process}. It is fixed to a Markov chain that incrementally adds Gaussian noise to each sample $\mathbf{x}_0$ in $T$ time-steps, producing a sequence of noisy samples $\mathbf{x}_1,\ldots,\mathbf{x}_T$ as latents of the same dimensionality as the original sample $\mathbf{x}_0$. Given $\mathbf{x}_0$ and a variance schedule $\beta_1,\ldots,\beta_T$ (\ie, how much noise is added at each time-step),
each $\mathbf{x}_t$ adheres to the following noisy sample distribution:
\begin{equation}
    q(\mathbf{x}_t | \mathbf{x}_0) = \mathcal{N} (\mathbf{x}_t; \sqrt{\bar{\alpha}_t} \mathbf{x}_0, (1-\bar{\alpha}_t) \mathbf{I}), \;\;\;\; \textrm{where} \;\; \alpha_t:=1-\beta_t, \; \bar{\alpha}_t:=\prod_{s=1}^t \alpha_s.
    \label{eq:2}
\end{equation}

\noindent\textbf{Reverse Process}. It corresponds to a learned Gaussian transition $p_\theta(\mathbf{x}_{t-1} | \mathbf{x}_t)$ parameterized by $\theta$, starting at $p(\mathbf{x}_T):=\mathcal{N}(\mathbf{x}_T;\mathbf{0},\mathbf{I})$. Each $p_\theta(\mathbf{x}_{t-1} | \mathbf{x}_t)$ is computed in two steps: 1) Reconstruct $\mathbf{x}_0$ from $\mathbf{x}_t$ with $u_\theta(\mathbf{x}_t, t)$, where $u_\theta$ is a learnable U-Net~\cite{ronneberger2015u}. 2) Compute $q\left( \mathbf{x}_{t-1} | \mathbf{x}_t, u_\theta(\mathbf{x}_t, t) \right)$, which has a closed-form solution given in Appendix.
Training is performed by minimizing the reconstruction error made by $u_\theta(\mathbf{x}_t, t)$ at each time-step $t$:
\begin{equation}
    \mathcal{L}_{ DM } = \mathop{\mathbb{E}}_{t, \mathbf{x}_0, \epsilon} \left[ \frac{\bar{\alpha}_t}{1-\bar{\alpha}_t} \lVert \mathbf{x}_0 - u_\theta (\sqrt{\bar{\alpha}_t} \mathbf{x}_0 + \sqrt{1-\bar{\alpha}_t}\epsilon , t) \rVert^2 \right],
    \label{eq:3}
\end{equation}
where $\epsilon \sim \mathcal{N}(\mathbf{0}, \mathbf{I})$ is a random noise. Note that we formulate the U-Net to predict $\mathbf{x}_0$ (equivalent to the $\epsilon$-formulation in ~\cite{ho2020denoising}).
As DM has no explicit feature encoder, we learn one to capture the modular attributes by leveraging the inductive bias in the diffusion time-steps.

\section{Method}
\label{sec:4}

\subsection{Theory}
\label{sec:4.1}

We reveal that in the forward diffusion process, fine-grained attributes are lost at an earlier time-step compared to coarse-grained ones, where the granularity is defined by the pixel-level changes when altering an attribute.
We first formalize the notion of attribute loss, \ie, when DM fails to distinguish the noisy samples drawn from two overlapping noisy sample distributions $q(\mathbf{x}_t|\mathbf{x}_0)$ and $q(\mathbf{y}_t|\mathbf{y}_0)$.

\noindent\textbf{Definition}. (Attribute Loss) \textit{Given a $\theta$-parameterized DM optimized by Eq.~\ref{eq:3}, we say that attribute $\mathcal{Z}_i$ is lost with degree $\tau$ at $t$ when $\mathop{\mathbb{E}}_{\mathbf{x}_0 \in \mathcal{X}} \left[ \mathrm{Err}(\mathbf{x}_0, \mathbf{y}_0=g_i \cdot \mathbf{x}_0, t) \right] \geq \tau$, where $\mathrm{Err}(\mathbf{x}_0, \mathbf{y}_0, t):=$}
\begin{equation}
     \frac{1}{2} \left[ \mathop{\mathbb{E}}_{q(\mathbf{x}_t|\mathbf{x}_0)} \left[ \mathbbm{1}\left( \lVert \hat{\mathbf{x}}_0 - \mathbf{x}_0 \rVert > \lVert \hat{\mathbf{x}}_0 - \mathbf{y}_0 \rVert \right) \right]+ \mathop{\mathbb{E}}_{q(\mathbf{y}_t|\mathbf{y}_0)} \left[\mathbbm{1}\left( \lVert \hat{\mathbf{y}}_0 - \mathbf{x}_0 \rVert < \lVert \hat{\mathbf{y}}_0 - \mathbf{y}_0 \rVert \right) \right] \right],
\end{equation}
\textit{with $\mathbbm{1}(\cdot)$ denoting the indicator function and $\hat{\mathbf{x}}_0=u_\theta(\mathbf{x}_t),\hat{\mathbf{y}}_0=u_\theta(\mathbf{y}_t)$ denoting the DM reconstructed samples from $\mathbf{x}_t,\mathbf{y}_t$, respectively.}

Intuitively, $\mathrm{Err}(\mathbf{x}_0, \mathbf{y}_0, t)$ measures attribute loss degree by how likely a DM falsely reconstructs $\mathbf{x}_t$ drawn from $q(\mathbf{x}_t|\mathbf{x}_0)$ closer to $\mathbf{y}_0$ instead of $\mathbf{x}_0$ (vice versa).
Hence, when $\mathbf{x}_0$ and $\mathbf{y}_0$ differ in attribute $\mathbf{z}_i$ modified by $g_i$, a larger $\mathrm{Err}(\mathbf{x}_0, \mathbf{y}_0, t)$ means that the attribute is more likely lost.

\noindent\textbf{Theorem}. (Attribute Loss and Time-step) \textit{1) For each $\mathcal{Z}_i$, there exists a smallest time-step $t(\mathcal{Z}_i)$, such that $\mathcal{Z}_i$ is lost with degree $\tau$ at each $t \in \{t(\mathcal{Z}_i),\ldots, T\}$. 2) $\exists \{\beta_i\}_{i=1}^T$ such that $t(\mathcal{Z}_i) > t(\mathcal{Z}_j)$ whenever $\lVert \mathbf{x}_0 - g_i \cdot \mathbf{x}_0 \rVert$ is first-order stochastic dominant over $\lVert \mathbf{x}_0 - g_j \cdot \mathbf{x}_0 \rVert$ with $\mathbf{x}_0 \sim \mathcal{X}$ uniformly.}

Intuitively, the first part of the theorem states that a lost attribute will not be regained as time-step $t$ increases, and there is a time-step $t(\mathcal{Z}_i)$ when $\mathcal{Z}_i$ becomes lost (with degree $tau$) for the first time.
The second part states that when $\lVert \mathbf{x}_0 - g_i \cdot \mathbf{x}_0 \rVert$ is more likely to take on a larger value than $\lVert \mathbf{x}_0 - g_j \cdot \mathbf{x}_0 \rVert$, the attribute $\mathcal{Z}_i$ (\ie, a coarse-grained attribute) is lost at a larger time-step compared to $\mathcal{Z}_j$ (\ie, a fine-grained attribute). We have the following proof sketch:

\noindent\textit{1) Increasing $t$ induces further attribute loss.} We show in Appendix that
$\mathrm{Err}(\mathbf{x}_0, \mathbf{y}_0, t) = \frac{1}{2} \mathrm{OVL}\left(q(\mathbf{x}_t | \mathbf{x}_0), q(\mathbf{y}_t | \mathbf{y}_0) \right)$ for an optimal DM, where $\mathrm{OVL}$ is the overlapping coefficient~\cite{inman1989overlapping}, \ie, overlapping area of probability density functions (PDFs). In particular,
\begin{equation}
    \mathrm{Err}(\mathbf{x}_0, \mathbf{y}_0, t) = \frac{1}{2} \mathrm{OVL}\left(q(\mathbf{x}_t | \mathbf{x}_0), q(\mathbf{y}_t | \mathbf{y}_0) \right) = \frac{1}{2} \left[ 1 - \mathrm{erf}\left(\frac{\lVert \sqrt{\bar{\alpha}_t} (\mathbf{y}_0-\mathbf{x}_0) \rVert}{2\sqrt{2(1-\bar{\alpha}_t)}}\right) \right].
    \label{eq:5}
\end{equation}
As $\bar{\alpha}_t$ decreases with an increasing $t$ from Eq.~\ref{eq:2}, and the error function $\mathrm{erf}(\cdot)$ is strictly increasing, $\mathrm{Err}(\mathbf{x}_0, \mathbf{y}_0, t)$ is strictly increasing in $t$ given any $\mathbf{x}_0,\mathbf{y}_0$. Hence $\mathop{\mathbb{E}}_{\mathbf{x}_0 \in \mathcal{X}} \left[ \mathrm{Err}(\mathbf{x}_0, \mathbf{y}_0=g_i \cdot \mathbf{x}_0, t) \right] \geq \mathop{\mathbb{E}}_{\mathbf{x}_0 \in \mathcal{X}} \left[ \mathrm{Err}(\mathbf{x}_0, \mathbf{y}_0=g_i \cdot \mathbf{x}_0, t(\mathcal{Z}_i)) \right]$ for every $t\geq t(\mathcal{Z}_i)$, which completes the proof.

\noindent\textit{2) Coarse-grained attributes are lost at a larger $t$.} Given that $\mathrm{erf}(\cdot)$ is strictly increasing and $\lVert \mathbf{x}_0 - g_i \cdot \mathbf{x}_0 \rVert$ is first-order stochastic dominant over $\lVert \mathbf{x}_0 - g_j \cdot \mathbf{x}_0 \rVert$, we have $\mathop{\mathbb{E}}_{\mathbf{x}_0 \in \mathcal{X}} \left[ \mathrm{Err}(\mathbf{x}_0, g_i \cdot \mathbf{x}_0, t) \right] > \mathop{\mathbb{E}}_{\mathbf{x}_0 \in \mathcal{X}} \left[ \mathrm{Err}(\mathbf{x}_0, g_j \cdot \mathbf{x}_0, t) \right]$ at every time-step $t$ using Eq.~\ref{eq:5}. Hence $t(\mathcal{Z}_i) > t(\mathcal{Z}_j)$ under any variance schedule $\{\beta_i\}_{i=1}^T$ such that $\mathcal{Z}_i$ is \emph{not} lost at $t(\mathcal{Z}_j)$, completing the proof.
Note that in practice, DM leverages a large $T$ with small $\{\beta_i\}_{i=1}^T$. This ensures that $\mathrm{Err}(\mathbf{x}_0, \mathbf{y}_0, t)$ slows increases with $t$ according to Eq.~\ref{eq:5}. Hence empirically, this theorem tends to hold.

\begin{figure}[t!]
    \centering
    \footnotesize
    \includegraphics[width=1.0\linewidth]{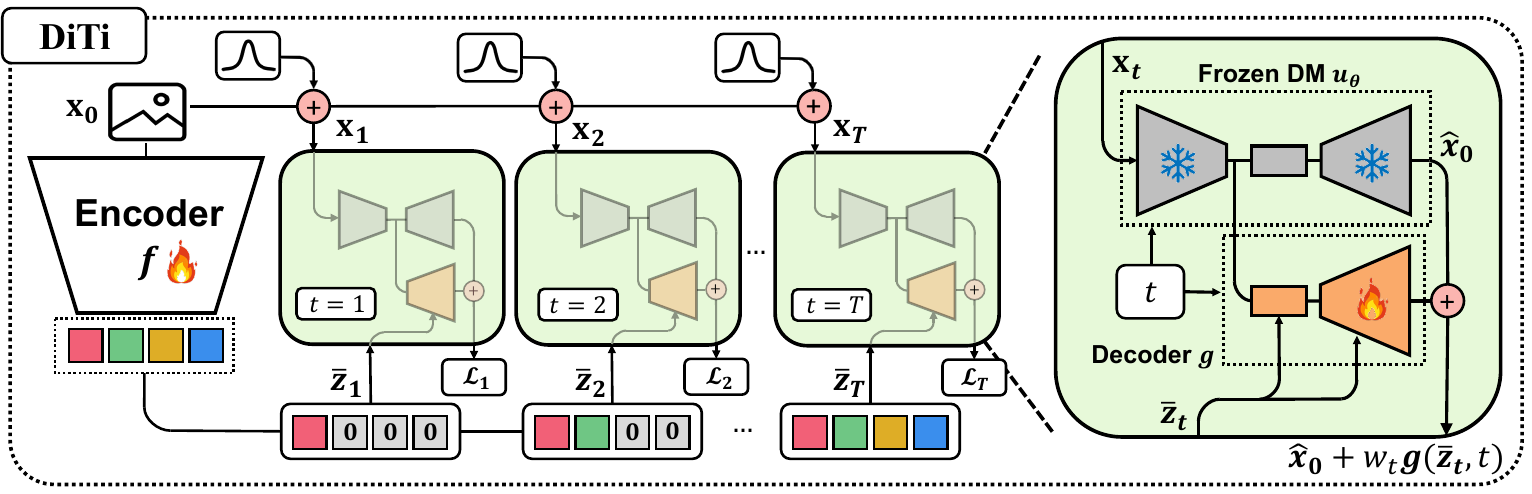}
    \caption{Illustration of our DiTi. We break down Eq.~\ref{eq:6} at each time-step. On the right, we show the detailed network design, where $\hat{\mathbf{x}}_0$ denotes the reconstructed $\mathbf{x}_0$ by the pre-trained DM.}
    \label{fig:3}
    \vspace*{-2mm}
\end{figure}

\subsection{Proposed Approach}
\label{sec:4.2}

The previous theoretical analysis leads to two interesting groundings:
1) Attribute loss is linked with the error of reconstructing $\mathbf{x}_0$ from the noisy $\mathbf{x}_t$.
2) Increasing time-step $t$ in forward diffusion process causes an expanding cumulative loss of attributes, with fine-grained attributes lost at a smaller $t$ and coarse-grained ones lost at a larger $t$.
They serve as an effective inductive bias for unsupervised learning:
1) making up for the reconstruction error can retrieve lost attributes,
2) an expanding set of features captures the expanding cumulatively lost attributes.
This motivates our approach below.

As illustrated in Figure~\ref{fig:3}, our model includes a frozen DM U-Net $u_\theta$ pre-trained on $\mathcal{D}$ using Eq.~\ref{eq:3} until convergence, a trainable encoder $f$ (trained to become a disentangled representation) that maps each image (without noise) to a $d$-dimensional feature $\mathbf{z} \in \mathbb{R}^d$, and a trainable decoder $g$ that maps $\mathbf{z}$ back to image space given $t$. In training, given $\mathbf{z}=f(\mathbf{x}_0)$, we partition it into $T$ disjoint subsets $\{\mathbf{z}_i\}_{i=1}^T$. At each time-step $t$, we construct a new feature $\bar{\mathbf{z}}_t = [\mathbf{z}_1,\ldots,\mathbf{z}_t, \mathbf{0}, \ldots, \mathbf{0}]$, which is only about $\{\mathbf{z}_i\}_{i=1}^t$ by masking $\mathbf{z}_{t+1},\ldots,\mathbf{z}_{T}$ as $\mathbf{0}$. Our training loss requires $\bar{\mathbf{z}}_t$ to compensate for the reconstruction error made by $u_\theta$ at each time-step $t$, given by:
\begin{equation}
    \mathcal{L} = \mathop{\mathbb{E}}_{t, \mathbf{x}_0, \epsilon} \; \underbrace{\left[\lambda_t \lVert \mathbf{x}_0 - \left( u_\theta (\sqrt{\bar{\alpha}_t} \mathbf{x}_0 + \sqrt{1-\bar{\alpha}_t}\epsilon , t) + w_t g(\bar{\mathbf{z}}_t, t) \right) \rVert^2 \right]}_\text{$\mathcal{L}_t$} ,
    \label{eq:6}
\end{equation}
where $\lambda_t, w_t$ are time-step weight and compensate strength. We follow PDAE~\cite{zhang2022unsupervised} to use fixed values computed from the variance schedule $\{\beta_t\}_{t=1}^T$ (details in Appendix).

To see the effect of Eq.~\ref{eq:6}, consider $t=1$, only $\mathbf{z}_1$ is used to compensate for reconstruction error, hence $\mathbf{z}_1$ effectively captures a (fine-grained) modular attribute $\mathcal{Z}_i$ with $t(\mathcal{Z}_i) = 1$ (lost at $t=1$) (such $\mathcal{Z}_i$ always exists for some threshold $\tau$ in our definition).
At $t=2$, the cumulatively lost attributes correspond to $\mathcal{Z}_i$ and $\mathcal{Z}_j$, where $t(\mathcal{Z}_j)=2$. As $\mathbf{z}_1$ already captures the attribute $\mathcal{Z}_i$, $\mathbf{z}_2$ is encouraged to further learn the attribute $\mathcal{Z}_j$ to minimize the reconstruction error at $t=2$.
Eventually with a large enough $T$, $\mathbf{x}_T$ becomes pure noise and all attributes are lost. By the above induction, $\{\mathbf{z}_i\}_{i=1}^T$ captures all the hidden attributes as a modular feature.
In contrast, the most related work PDAE differs from Eq.~\ref{eq:6} by replacing $g(\bar{\mathbf{z}}_t)$ with $g(\bar{\mathbf{z}})$.
This means that PDAE uses the full-dimensional feature $\mathbf{z}$ to compensate the reconstruction error at all time-steps, which fails to leverage the inductive bias discussed in Section~\ref{sec:4.1}. We show in Section~\ref{sec:5} that this simple change leads to drastic differences in both inference and generation tasks.

\noindent\textbf{Implementation Details}. We highlight some design considerations with ablations in Section~\ref{sec:5.2}.
\begin{itemize}[leftmargin=+0.25in,itemsep=1pt,topsep=0pt,parsep=0pt]
    \item Number of Subsets $k$. In practice, the number of subsets for partitioning $\mathbf{z}$ can be smaller than $T$, \eg, we use feature dimension $d=512$ that is smaller than $T=1,000$. This means that adjacent time-steps may share a modular feature $\mathbf{z}_i$. We compare the choice of $k$ later.
    \item Partition Strategy. The simplest way is to use the same number of feature dimensions for all modular features $\mathbf{z}_i$ (\ie, balanced). However, we empirically find that features corresponding to time-step 100-300 requires more dimensions to improve convergence of Eq.~\ref{eq:6}. We design an imbalanced partition strategy and compare it with the balanced one.
    \item Optimization Strategy. In training, we experiment detaching the gradient of $\mathbf{z}_1,\ldots,\mathbf{z}_{t-1}$ to form $\bar{\mathbf{z}}_t$, \ie, training only $\mathbf{z}_t$ at time-step $t$. We find that it leads to improved disentanglement quality at the cost of slower convergence.
\end{itemize}

\section{Experiments}
\label{sec:5}

\subsection{Settings}
\label{sec:5.1}

\noindent\textbf{Datasets}. We choose real-world datasets to validate if DiTi learns a {disentangled representation} of the generative attributes: 1) Celebrity Faces Attributes (CelebA)~\cite{liu2015faceattributes} is a large-scale face attributes dataset. Each face image has 40 binary attribute labels, delineating facial features and characteristics, such as expressions, accessories, and lighting conditions. 2) Flickr-Faces-HQ (FFHQ)~\cite{karras2019style} contains 70,000 high-quality face images obtained from Flickr. 3) We additionally used the Labeled Faces in the Wild (LFW) dataset~\cite{Huang2007a} that provides continuous attribute labels. 4) Bedroom is part of the Large-scale Scene UNderstanding (LSUN) dataset~\cite{yu2015lsun} that contains around 3 million images. We apply off-the-shelf image classifiers following \cite{yang2021semantic} to determine scene attribute labels.

\noindent\textbf{Evaluation Protocols}. Our evaluation is based on the properties of a disentangled representation (Section~\ref{sec:3.1}). For \textbf{inference}, we perform unsupervised learning on CelebA train split or FFHQ and test the linear-probe attribute prediction accuracy on CelebA test split, measured by Average Precision (AP). We also evaluate the challenging attribute regression task using LFW, where the metrics are Pearson correlation coefficient (Pearson's r) and Mean Squared Error (MSE). For \textbf{counterfactual generation}, we extend the conventional interpolation and manipulation technique~\cite{preechakul2022diffusion} to tailor for our needs: 1) Instead of interpolating the whole features of an image pair, we interpolate only a feature subset while keeping its complement fixed. With a disentangled representation, only the attribute captured by the subset will change in the generated counterfactuals, \eg, changing only expression between two faces. 2) In manipulation, an image is edited by altering its feature along the direction of an attribute classifier weight, where the classifier is trained on whole features. We constrain the classifier by ProbMask~\cite{zhou2021effective} to use a small subset of the most discriminative feature dimensions (32 out of 512 dimensions). For a disentangled representation, this should have minimal impact on the generation quality, as each single attribute is encoded in a compact feature subset. Detailed algorithms are in Appendix.

\noindent\textbf{Baselines} include 3 representative lineups: 1) Conventional unsupervised disentanglement methods $\beta$-TCVAE~\cite{chen2018isolating} and IB-GAN~\cite{jeon2021ib}; 2) Self-supervised learning methods SimCLR~\cite{chen2020simple}; 3) The most related works Diff-AE~\cite{preechakul2022diffusion} and PDAE~\cite{zhang2022unsupervised} that also learn a representation by making up DM's reconstruction errors.

\noindent\textbf{Implementation Details}. We followed the network design of encoder $f$ and decoder $g$ in PDAE and adopted its hyper-parameter settings (\eg, $\lambda_t,w_t$ in Eq.~\ref{eq:5}, details in Appendix). This ensures that any emerged property of disentangled representation is solely from our leverage of the inductive bias in Section~\ref{sec:4.1}. We also used the same training iterations as PDAE, \ie, 290k iterations on CelebA, 500k iterations on FFHQ, and 540k iterations on Bedroom. Hence our DiTi training is as efficient as PDAE. Our experiments were performed on 4 NVIDIA A100 GPUs.

\subsection{Inference}
\label{sec:5.2}

\begin{table}
    \centering
    \captionsetup{font=footnotesize,labelfont=footnotesize,skip=5pt}
    \setlength\tabcolsep{5pt}
    \scalebox{0.8}{
        \def\arraystretch{1.1}
        \begin{tabular}{p{4.6cm} p{1cm}<{\centering} p{2.0cm}<{\centering} p{1.2cm}<{\centering} p{0.5cm} p{1cm}<{\centering} p{2.0cm}<{\centering} p{1.2cm}<{\centering}}
        \hline\hline
                \multicolumn{1}{c}{\multirow{2}{*}{\large{\textbf{Method}}}} & \multicolumn{3}{c}{\textbf{CelebA}} & & \multicolumn{3}{c}{\textbf{FFHQ}}\\ \cmidrule(lr){2-4}\cmidrule(lr){6-8}
                \multicolumn{1}{c}{} & {AP}\ $\uparrow$ & {Pearson's r}\ $\uparrow$ & {MSE}\ $\downarrow$ &  & {AP}\ $\uparrow$ & {Pearson's r}\ $\uparrow$ & {MSE}\ $\downarrow$ \\
                \hline
                $\beta$-TCVAE~\cite{chen2018isolating} & 45.0  & 0.378  & 0.573 & & 43.2  & 0.335  & 0.608   \\
                IB-GAN~\cite{jeon2021ib} & 44.2  & 0.307  & 0.597  & & 42.8  & 0.260  & 0.644  \\
                Diff-AE~\cite{preechakul2022diffusion} & 60.3  & 0.598  & 0.421 & & 60.5  & 0.606  & 0.410   \\
                PDAE~\cite{zhang2022unsupervised} & 60.2  & 0.596  & 0.410  & & 59.7 & 0.603 & 0.416  \\
                SimCLR~\cite{chen2020simple}& 59.7 & 0.474 & 0.603 & & 60.8 & 0.481 & 0.638   \\
                \textbf{DiTi (Ours)} & \cellcolor{mygray} \textbf{62.3} & \cellcolor{mygray} \textbf{0.617} & \cellcolor{mygray} \textbf{0.392}  & \cellcolor{mygray} & \cellcolor{mygray} \textbf{61.4} & \cellcolor{mygray} \textbf{0.622} & \cellcolor{mygray} \textbf{0.384}   \\
        \hline\hline
        \end{tabular}
    }
    \caption{AP (\%) on CelebA attribute classification and Pearson's r, MSE on LFW attribute regression. The first column shows the training dataset.}
    \label{tab:1}
    \vspace{-2mm}
\end{table}
\begin{figure}[t!]
    \centering
    \captionsetup{font=footnotesize,labelfont=footnotesize,skip=5pt}
    \includegraphics[width=1.0\linewidth]{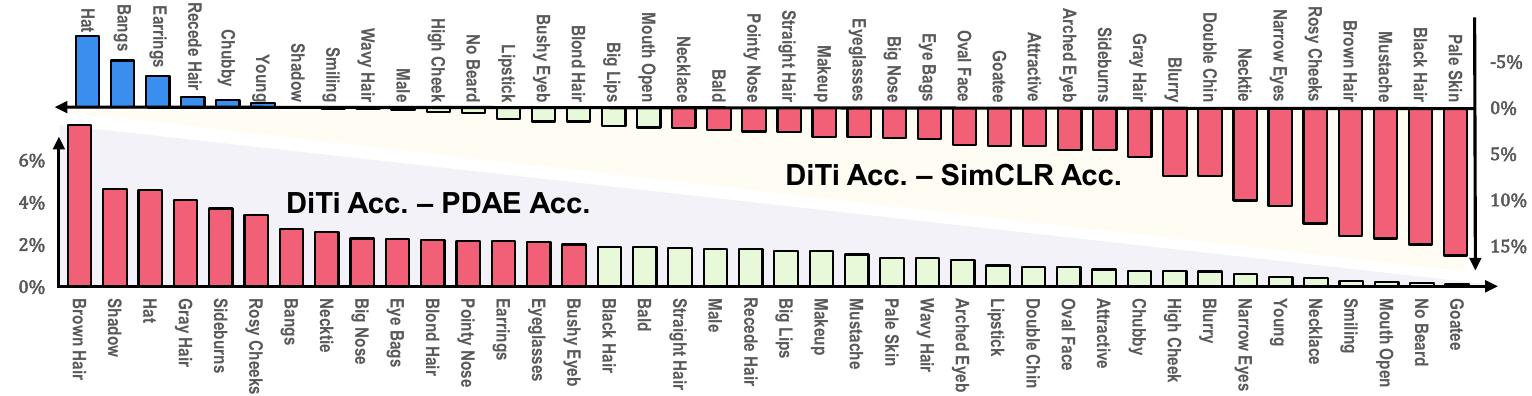}
    \caption{Improvements in attribute classification precision of our DiTi over PDAE (bottom) and over SimCLR (top). Improvements more than 2\% are highlighted with red bars. Negative values are marked with blue bars.}
    \label{fig:4}
    \vspace*{-2mm}
\end{figure}

\textbf{Comparison with Diff-AE and PDAE}. In Table~\ref{tab:1}, our proposed DiTi significantly outperforms the two related works. We break down the improved accuracy over PDAE on each attribute in Figure~\ref{fig:4} bottom, where DiTi is consistently better on all attributes. The result, together with the previous analysis of Figure~\ref{fig:manipulate}b, validates that DiTi exhibits the sample inference property of a disentangled representation. It also supports our analysis on the baselines' deficiency in Section~\ref{sec:2}, \ie, they may digress to use spurious correlation to supplement reconstruction error. For example, the most improved attribute is ``Brown Hair'', which strongly correlates with the female gender.

\textbf{Comparison with SimCLR}. SimCLR does not perform well in attribute inference, surpassed even by PDAE.
This is because SimCLR learns a representation invariant to augmentations (\eg, color jittering), which disregards its related attributes, such as hair or skin color. This is verified by the large improvements of DiTi over SimCLR on ``Pale Skin'' accuracy in Figure~\ref{fig:4} top.
The deficiency cannot be addressed by simply removing augmentations, which leads to severe performance degradation (see Appendix).
Note that SimCLR slightly outperforms DiTi on some unnoticeable attributes (\eg, earrings). This suggests that some fine-grained attributes are not well disentangled. To tackle this, we will improve model expressiveness (\eg, using Stable Diffusion~\cite{rombach2022high} as the frozen DM) and explore more advanced training objectives~\cite{song2023consistency} as future work.

\begin{wrapfigure}{r}{0.55\textwidth}
    \centering
    \vspace{-4mm}
    \captionsetup{font=footnotesize,labelfont=footnotesize,skip=5pt}
        \scalebox{0.8}{
            \begin{tabular}{p{1.3cm}<{\centering} p{1.1cm}<{\centering} p{0.6cm}<{\centering} p{0.9cm}<{\centering} p{2cm}<{\centering} p{1.1cm}<{\centering}}
                \hline\hline
                \textbf{Imbalance} & \textbf{Detach} & \textbf{$k$} & \textbf{AP}\ $\uparrow$ & \textbf{Pearson's r}\ $\uparrow$ & \textbf{MSE}\ $\downarrow$ \\
                \hline
                \checkmark & &  16 & 62.1  & 0.619  &  0.389     \\
                \checkmark & &  32 & 62.0 & 0.615 & 0.392     \\
                \checkmark & &  64 & 62.3 & 0.617 & 0.392     \\
                \checkmark & &  128 & 61.9 & 0.616 & 0.391     \\
                &  &  64 & 61.9 & 0.604 &  0.410   \\
                & \checkmark &  64 & 62.5 & 0.590 & 0.422    \\
              \checkmark &\checkmark & 64 & 62.4 & 0.604 &0.405 \\
                \hline \hline
            \end{tabular}
        }
    \captionof{table}{Ablations on DiTi designs on CelebA. \textit{Imbalance} and \textit{Detach}: using our partition and optimization strategy.}
    \label{tab:2}
\vspace*{-3mm}
\end{wrapfigure}
\textbf{Comparison with $\beta$-TCVAE and IB-GAN}. It is perhaps not surprising that existing unsupervised disentanglement methods perform poorly. Their previous successes are limited to synthetic datasets, and their models are far less expressive compared to DM. In fact, their pursuit of a disentangled representation by information bottleneck proves to be ever-elusive~\cite{locatello2019challenging}.

\textbf{Ablation on \#Subsets $k$}. From Table~\ref{tab:2} first 4 lines, we observe that inference results are not sensitive to the subset number $k$ (details in Section~\ref{sec:4.2}). We choose $k=64$ with the highest AP. This means that our model can learn up to 64 hidden attributes, whose combinations are certainly diverse enough for our chosen datasets.

\textbf{Ablation on Partition \& Optimization Strategy}. We devise an imbalanced partition strategy to allocate more feature dimensions for $\mathbf{z}_i$ corresponding to $t\in \{100,\ldots,300\}$, as this time-step range contributes the most to the overall loss (Figure A9). This leads to faster convergence and improved performance in Table~\ref{tab:2} (line 3 vs. line 5). For optimization strategy, we tried detaching the gradient of $\mathbf{z}_1,\ldots,\mathbf{z}_{t-1}$ at time-step $t$ to prevent them from capturing the lost attribute at $t$. Yet it leads to slower convergence, which especially hurts the challenging regression task. We leave it as future work to explore more principled ways of improving feature modularity.

\begin{figure}[t!]
    \centering
    \captionsetup{font=footnotesize,labelfont=footnotesize,skip=5pt}
    \includegraphics[width=1.0\linewidth]{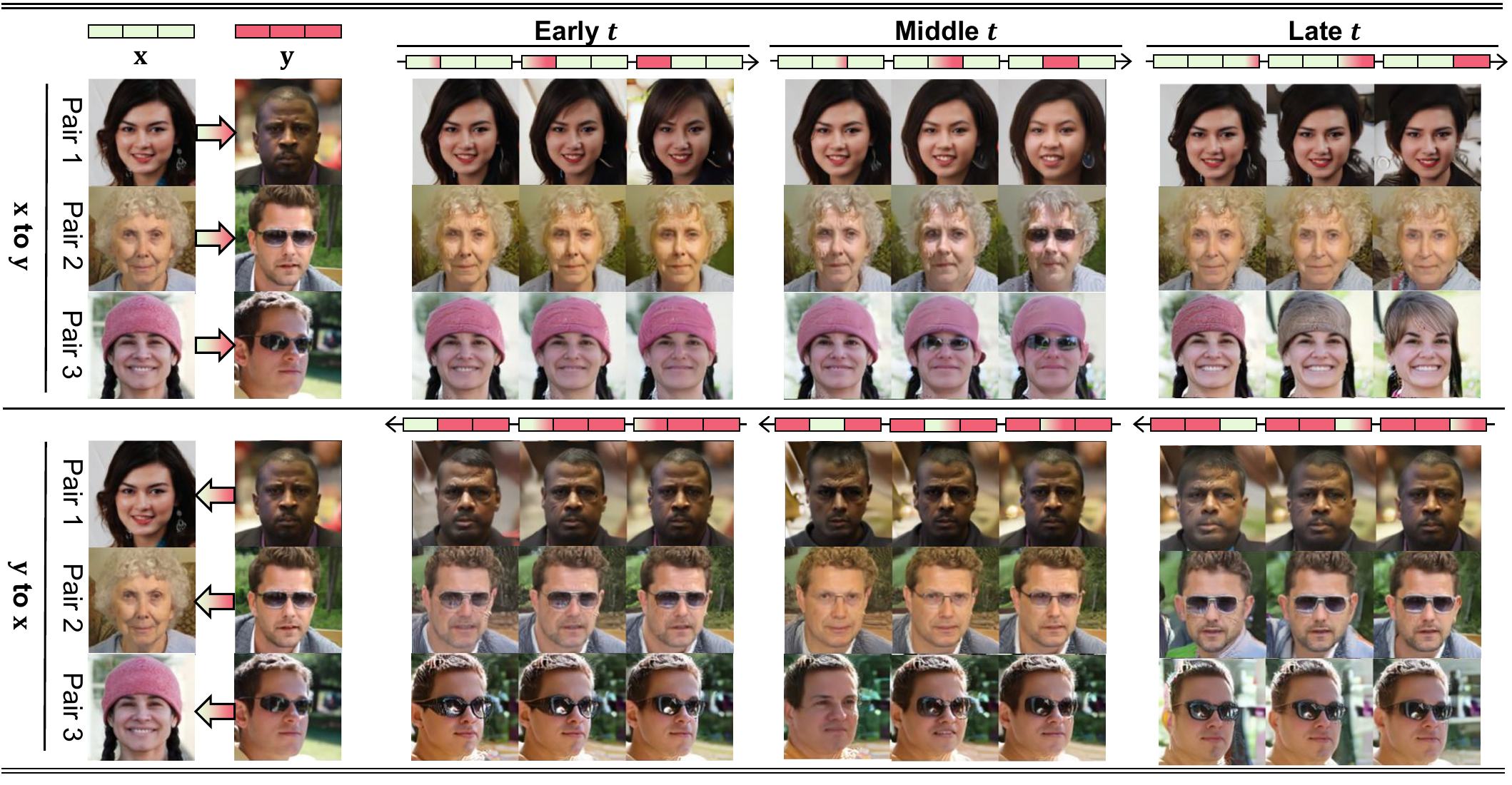}
    \caption{Counterfactual generations on FFHQ by interpolating a feature subset of 3 image pairs $(\mathbf{x},\mathbf{y})$. We partition $\{\mathbf{z}_i\}_{i=1}^T$ according to early $t\in (0, T/3]$, middle $t\in (T/3, 2T/3]$ and late $t\in (2T/3, T]$. For each subset, we show results in 2 directions ($\mathbf{x}\to\mathbf{y}$, $\mathbf{y}\to \mathbf{x}$) under 3 interpolation scales (color gradient$\uparrow$, scale$\uparrow$).}
    \label{fig:5}
    \vspace*{-5mm}
\end{figure}

\vspace{-3mm}
\subsection{Counterfactual Generation}
\label{sec:5.3}

\begin{figure}[t!]
    \centering
    \captionsetup{font=footnotesize,labelfont=footnotesize,skip=5pt}
    \includegraphics[width=1.0\linewidth]{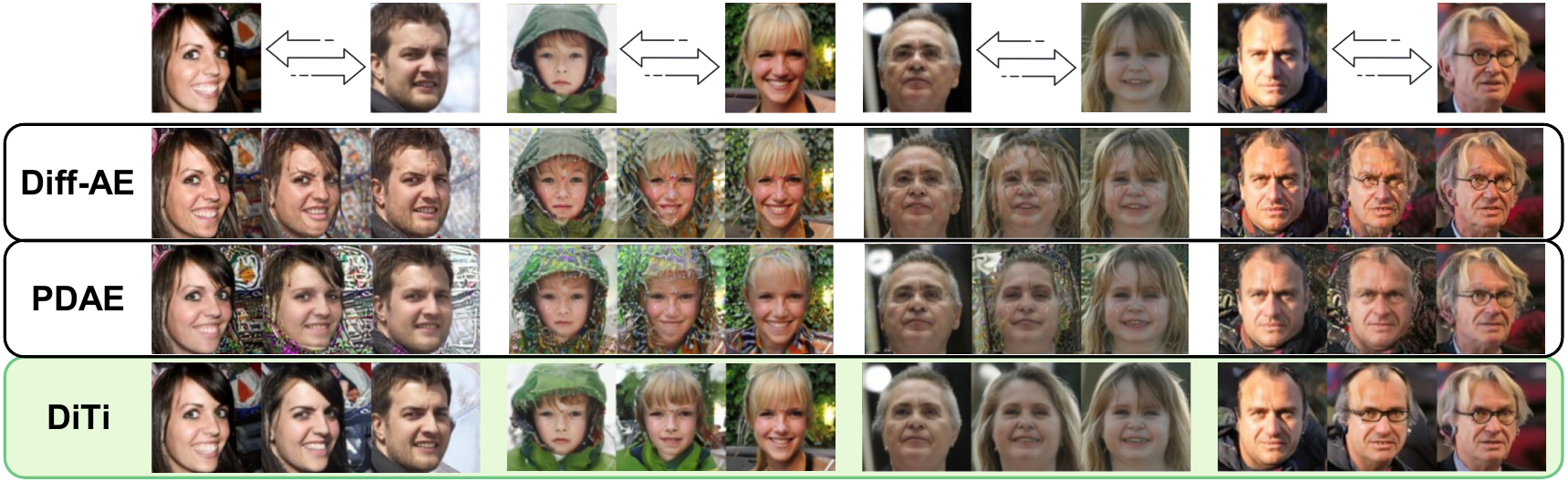}
    \caption{Results of interpolating the whole feature on FFHQ. Baselines have distortions during interpolation.}
    \label{fig:7}
    \vspace*{-4mm}
\end{figure}

\textbf{FFHQ Results}.
Figure~\ref{fig:5} shows DiTi generations by interpolating a feature subset while fixing its complement. Different from full-feature interpolation (Figure~\ref{fig:7}), there are two directions: $\mathbf{x}\to \mathbf{y}$ fixes the complement to the corresponding feature values of $\mathbf{x}$, and $\mathbf{y}\to \mathbf{x}$ fixes it to those of $\mathbf{y}$.
By identifying how images are changing during interpolation (left to right for top, right to left for bottom) and comparing the changes across the 3 columns, we have several observations:
\begin{enumerate}[leftmargin=+0.15in,itemsep=1pt,topsep=0pt,parsep=0pt]
    \item Interpolating different subsets modifies different attributes. For example, subset of early $t$ mainly controls micro-expressions, such as mouth corners, cheeks, or eye details. The interpolation hardly changes a person's identity, but certainly moves one's temperament towards the other person. In contrast, subset of middle $t$ is more responsible for identity-related attributes (\eg, proportion of facial features) and accessories (\eg, eyeglasses). Finally, late $t$ controls attributes such as hairstyle, pose, and even position of the face in the image (\eg, the bottom right face moves towards center).
    \item From early to late $t$, the corresponding feature subset controls a more coarse-grained feature, where the granularity is defined based on the pixel-level changes. For example, changing hairstyle or removing hat (subset of late $t$) modifies more pixels than changing facial feature proportions (subset of middle $t$). This is in line with our theoretical analysis in Section~\ref{sec:4.1}. Notably, both our feature subset partition and the correspondence between subsets and attributes granularity are known a priori, without relying on manual discovery techniques such as latent traversal.
    \item Our generated counterfactuals have minimal distortion or artifact, and demonstrate modular properties (\eg, subset of late $t$ controls pose with minimal impact on facial features and accessories). These are strong proofs that DiTi learns a disentangled representation~\cite{besserve2018counterfactuals}. We show in Appendix that all generative baselines do not exhibit these traits.
\end{enumerate}

\begin{figure}[t!]
    \centering
    \captionsetup{font=footnotesize,labelfont=footnotesize,skip=5pt}
    \includegraphics[width=1.0\linewidth]{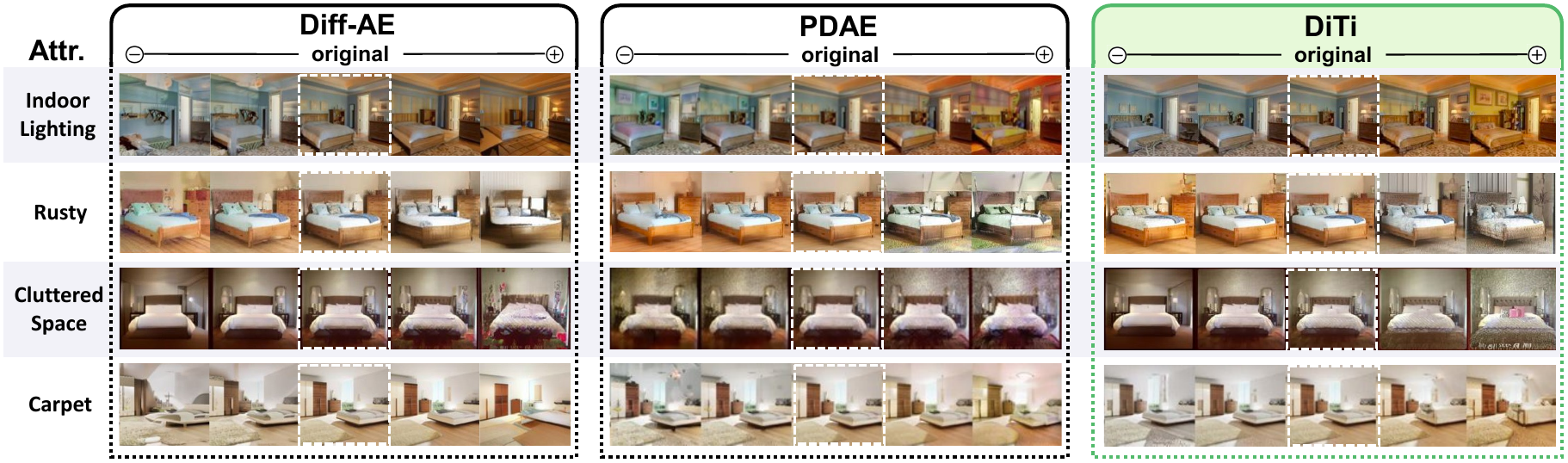}
    \caption{Counterfactual generations on Bedroom by manipulating 32 out of 512 feature dimensions.}
    \label{fig:6}
    \vspace*{-4mm}
\end{figure}

\textbf{Bedroom Manipulation}. Bedroom is a more challenging dataset, where room scenes are much more complex than facial images (\eg, various layouts, decorations, objects). As shown in Figure~\ref{fig:6}, DiTi is the only method that generates faithful counterfactuals. 1st row: only DiTi can alter lighting without making significant changes to room furniture. 2nd row: baselines fail to modify the rustic feeling without artifacts or changing room layouts. 3rd row: DiTi makes the room more cluttered by introducing additional objects. Last row: only DiTi adds an additional carpet.

\textbf{Whole-feature Interpolation}. Under this conventional setting, as shown in Figure~\ref{fig:7}, only DiTi smoothly interpolates between the attributes of each image pair, with minimal distortion and artifact along the trajectories. This means that our DiTi can generate faithful counterfactuals (\ie, valid images in the support of data distribution), which validates our disentanglement quality~\cite{besserve2018counterfactuals}. For example, if a feature only captures the attribute ``smile'' (\ie, disentangled), interpolating this feature between a smiling person and a non-smiling one leads to the gradual transition of expression. However, if a feature entangles multiple attributes, such interpolation will alter all of them or even cause distortion when the interpolated feature is out of the data distribution.

\vspace{-2mm}
\section{Conclusion}
\label{sec:6}
\vspace{-2mm}
We presented a novel unsupervised method to learn a disentangled representation, which leverages the inductive bias of diffusion time-steps.
In particular, we reveal an inherent connection between time-step and hidden modular attributes that generate data faithfully, enabling a simple and effective approach to disentangle the attributes by learning a time-step-specific feature.
The learned feature improves downstream inference and enables counterfactual generation, validating its disentanglement quality.
As future work, we will seek additional inductive bias to improve disentanglement, \eg, using text as a disentangled template by exploring text-to-image diffusion models, and devise practical optimization techniques to enable faster convergence.

\section{Acknowledgement}

This research is supported by Microsoft Research Asia, the National Research Foundation, Singapore under its AI Singapore Programme (AISG Award No: AISG2-RP-2021-022), MOE AcRF Tier 2 (MOE2019-T2-2-062), Wallenberg-NTU Presidential Postdoctoral Fellowship, the Lee Kong Chian (LKC) Fellowship fund awarded by Singapore Management University, and the DSO Research Grant (Fund Code MG22C03).

\bibliography{ref}
\bibliographystyle{iclr2024_conference}

\newpage

\appendix
\section{Appendix}

\renewcommand{\thetable}{A\arabic{table}}
\renewcommand{\thefigure}{A\arabic{figure}}
\renewcommand{\theequation}{A\arabic{equation}}

This is the Appendix for ``Exploring Diffusion Time-steps for Unsupervised Representation Learning''.
Table~\ref{tbl:notation} summarizes the abbreviations and the symbols used in the main paper.

This appendix is organized as follows:
\begin{itemize}
\item Section~\ref{sec:A} discusses the limitation and broader impact of our work.
\item Section~\ref{sec:B} provides additional details on the DM formulation and the hyper-parameter choice in Eq. (6).
\item Section~\ref{sec:C} gives the full proof to our Theorem, and shows the sufficiency of disentangled representation in minimizing Eq. (6).
\item Section~\ref{sec:D} presents the algorithm for training, counterfactual generation and manipulation, the network architecture and additional training details.
\item Section~\ref{sec:E} presents additional generation results supplementary to Figure~\ref{fig:manipulate}, Figure~\ref{fig:5}, Figure~\ref{fig:6} and Figure~\ref{fig:7}.
\end{itemize}

\begin{table}[htbp]
\centering
\begin{center}
\begin{tabular}{lll}
\toprule\toprule
Abbreviation/Symbol & Meaning\\
\midrule
\multicolumn{2}{c}{\underline{\emph{Abbreviation}}} \vspace{3pt} \\
DM     & Denoising Diffusion Probabilistic Model\\
OVL      & Overlapping Coefficient\\
AP     & Average Precision\\
MSE     & Mean Squared Error\\
slerp     & Spherical linear interpolation \\
lerp     & Linear interpolation \\
\midrule
\multicolumn{2}{c}{\underline{\emph{Symbol in Theory}}} \vspace{3pt} \\
$\mathcal{X}$     & Sample space \\
$\mathcal{Z}$     & Vector space\\
$\mathcal{Z}_i$     & Modular attribute\\
$g_i$         & Group element acting on $\mathcal{Z}_i$ \\
$\Phi$              & Injective mapping $\mathcal{Z}\to \mathcal{X}$\\
$f$     & Disentangled representation $\mathcal{X}\to \mathcal{Z}$\\
$\mathrm{erf}$          & Error function \\

\midrule
\multicolumn{2}{c}{\underline{\emph{Symbol in Algorithm}}} \vspace{3pt} \\
$\mathbf{x}_0$        & Original sample \\
$\mathbf{x}_t$      & Noisy sample after $t$ forward step\\
$q(\cdot)$       & Distribution in the encoding process \\
$p_\theta(\cdot)$        & Distribution in the $\theta$-parameterized decoding process \\
$\theta$            & Parameter of U-Net  \\
$u_\theta$        & $\theta$-parameterized U-Net \\
$\hat{\mathbf{x}}_0$         & Reconstructed $\mathbf{x}_0$ \\
$\mathbf{z}_i$       & $i$-th modular attribute value\\
$\bar{\mathbf{z}}_i$      & $[\mathbf{z}_1,\ldots,\mathbf{z}_i, 0,\ldots,0]$ \\
$T$        & Total time-steps \\
$\beta_1,\ldots,\beta_T$        & Variance schedule \\
$\alpha_t$        & $1-\beta_t$ \\
$\bar{\alpha}_t$        & $\prod_{s=1}^t \alpha_s$ \\

\bottomrule\bottomrule
\end{tabular}
\end{center}
\caption{List of abbreviations and symbols used in the paper.}
\label{tbl:notation}
\end{table}

\newpage
\section{Limitation and Broader Impact}
\label{sec:A}

\noindent\textbf{Limitation}. A disentangled representation is a sufficient condition to minimize our objective in Eq. (6), but not a necessary one. In particular, let $t(\mathcal{Z}_i) < t(\mathcal{Z}_j)$. At $t=t(\mathcal{Z}_i)$, attribute $\mathcal{Z}_i$ is lost with a larger degree compared to $\mathcal{Z}_j$. Hence while Eq. (6) trains $\mathbf{z}_t$ to mainly capture $\mathcal{Z}_i$, it may not fully remove $\mathcal{Z}_j$. 
As future work, we will try other methods to enforce the modularity between feature subsets, \eg, by using invariant learning as in~\cite{wang2021self}.

\noindent\textbf{Broader Impact}. While our learned model (encoder and decoder) can be used to generate synthetic data for malicious purposes, researchers have built models to predict fake content accurately. Moreover, the focus of our study is not improving the generation fidelity, but to learn disentangled representation, which leads to robust and fair AI models resilient to spurious correlations.
\section{Additional Details for Approach}
\label{sec:B}

\textbf{Closed Form of $q\left( \mathbf{x}_{t-1} | \mathbf{x}_t, u_\theta(\mathbf{x}_t, t) \right)$}. Given by $\mathcal{N}(\mathbf{x}_{t-1} | \tilde{\boldsymbol{\mu}}_{t}(\mathbf{x}_t,\mathbf{x}_0), \tilde{\beta}_t \mathbf{I} )$, where
\begin{equation}
    \tilde{\boldsymbol{\mu}}_{t}(\mathbf{x}_t,\mathbf{x}_0) = \frac{\sqrt{\bar{\alpha}_{t-1}} \beta_t}{1-\bar{\alpha}_t} \mathbf{x}_0 + \frac{\sqrt{\alpha_t} (1-\bar{\alpha}_{t-1})}{1-\bar{\alpha}_t} \mathbf{x}_t, \,\, \tilde{\beta}_t = \frac{1-\bar{\alpha}_{t-1}}{1-\bar{\alpha}_t} \beta_t.
\end{equation}

\textbf{Equivalent Formulation}. The simplified objective in DDPM~\cite{ho2020denoising} is given by:
\begin{equation}
    \mathcal{L}_{ DM } = \mathop{\mathbb{E}}_{t, \mathbf{x}_0, \epsilon}  \lVert \epsilon - \epsilon_\theta (\sqrt{\bar{\alpha}_t} \mathbf{x}_0 + \sqrt{1-\bar{\alpha}_t}\epsilon , t) \rVert^2,
    \label{eq:simple}
\end{equation}
where $\epsilon_\theta$ is a $\theta$-parameterized U-Net~\cite{ronneberger2015u} to predict the added noise. From Eq. (2), we have
\begin{equation}
    \epsilon = \frac{\mathbf{x}_t-\sqrt{\bar{\alpha}_t} \mathbf{x}_0}{\sqrt{1-\bar{\alpha}_t}}, \,\, \epsilon_\theta = \frac{\mathbf{x}_t-\sqrt{\bar{\alpha}_t} u_\theta}{\sqrt{1-\bar{\alpha}_t}},
    \label{eq:equal}
\end{equation}
where we slightly abuse the notation to denote the reconstructed $\mathbf{x}_0$ from our U-Net as $u_\theta$. Taking Eq.~\ref{eq:equal} into Eq.~\ref{eq:simple} yields Eq. (3).

\textbf{Time-step Weight $\lambda_t$ and Compensate Strength $w_t$}. The PDAE objective is given by
\begin{equation}
    \mathcal{L}_{ PDAE } = \mathop{\mathbb{E}}_{t, \mathbf{x}_0, \epsilon} \left[ \lambda^p_t  \lVert \epsilon - \epsilon_\theta (\sqrt{\bar{\alpha}_t} \mathbf{x}_0 + \sqrt{1-\bar{\alpha}_t}\epsilon , t)  + w^p_t g(\mathbf{z},t)  \rVert^2 \right],
    \label{eq:pdae}
\end{equation}
where
\begin{equation}
    \lambda_t^p = \left( \frac{1}{1+\mathrm{SNR}(t)} \right)^{0.9} \left( \frac{\mathrm{SNR}(t)}{1+\mathrm{SNR}(t)} \right)^{0.1}, \mathrm{SNR}(t) = \frac{\bar{\alpha}_t}{1-\bar{\alpha}_t}, w_t^p=\frac{\sqrt{\alpha_t}(1-\bar{\alpha}_{t-1})}{\sqrt{1-\bar{\alpha}_t}}.
\end{equation}
Taking Eq.~\ref{eq:equal} into Eq.~\ref{eq:pdae} yields:
\begin{equation}
    \lambda_t = \frac{\bar{\alpha}_t}{1-\bar{\alpha}_t} \lambda_t^p, \,\, w_t = \frac{ \sqrt{1-\bar{\alpha}_t} }{ \sqrt{\bar{\alpha}_t} } w_t^p = \sqrt{ \frac{\alpha_t}{\bar{\alpha}_t} } (1-\bar{\alpha}_t).
\end{equation}
\section{Theory}
\label{sec:C}

\begin{wrapfigure}{r}{0.4\textwidth}
\vspace{-8mm}
\captionsetup{font=footnotesize,labelfont=footnotesize}
    \centering
    \includegraphics[width=.9\linewidth]{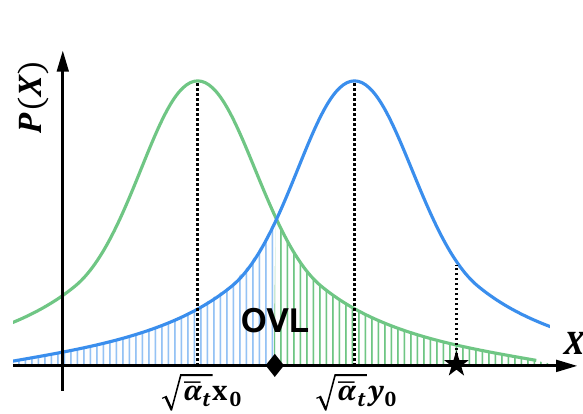}
    
    \caption{The PDFs of $q(\mathbf{x}_t|\mathbf{x}_0)$ (green) and $q(\mathbf{y}_t|\mathbf{y}_0)$ (blue). Without loss of generality, we consider the 1-D case of $y_0 > x_0$. Their means are computed from Eq. (2).}
    \label{fig:ovl}
    \vspace{-8mm}
\end{wrapfigure}
\textbf{Full Proof to the Theorem}. We list the theorem below for reference.

\textbf{Theorem}. (Attribute Loss and Time-step) \textit{1) For each $\mathcal{Z}_i$, there exists a smallest time-step $t(\mathcal{Z}_i)$, such that $\mathcal{Z}_i$ is lost with degree $\tau$ at each $t \in \{t(\mathcal{Z}_i),\ldots, T\}$. 2) $\exists \{\beta_i\}_{i=1}^T$ such that $t(\mathcal{Z}_i) > t(\mathcal{Z}_j)$ whenever $\lVert \mathbf{x}_0 - g_i \cdot \mathbf{x}_0 \rVert$ is first-order stochastic dominant over $\lVert \mathbf{x}_0 - g_j \cdot \mathbf{x}_0 \rVert$ with $\mathbf{x}_0 \sim \mathcal{X}$ uniformly.}

\textit{Proof}. We start by showing $\mathrm{Err}(\mathbf{x}_0, \mathbf{y}_0, t) = \frac{1}{2} \mathrm{OVL}\left(q(\mathbf{x}_t | \mathbf{x}_0), q(\mathbf{y}_t | \mathbf{y}_0) \right)$. Without loss of generality, we show a 1-D sample space $\mathcal{X}$ in Figure~\ref{fig:ovl}.
The minimum $\mathrm{Err}_\theta$ is obtained when given each noisy sample $\mathbf{x}$, DM reconstructs towards $\mathbf{x}_0$ if $q(\mathbf{x}|\mathbf{x}_0) > q(\mathbf{x}|\mathbf{y}_0)$ and vice versa for $\mathbf{y}_0$, \eg, reconstructing $\star$ as $\mathbf{y}_0$.
However, this maximum likelihood estimation fails when a noisy sample is drawn from $q(\mathbf{x}_t | \mathbf{x}_0)$ (green PDF), but with a value larger than the intersection point of the two PDFs ($\blacklozenge$), and similar arguments go for $q(\mathbf{y}_t | \mathbf{y}_0)$ (blue PDF).
The error rate caused by the two failure cases corresponds to the green shaded area and blue one, respectively, leading to an average $\mathrm{Err}_\theta$ of $\frac{1}{2}$ of the OVL.

To compute the OVL, it is trivial in the 1-D case by leveraging the Cumulative Distribution Function (CDF) of Gaussian distribution. Given that the two distributions have equal variance from Eq. (2), the intersection point is given by $\frac{\sqrt{\bar{\alpha}_t} \mathbf{x}_0 + \sqrt{\bar{\alpha}_t} \mathbf{y}_0 }{2}$. For a Gaussian distribution $\mathcal{N}(\mu, \sigma^2)$, its CDF is given by $\frac{1}{2} \left[  1+ \mathrm{erf}(\frac{x-\mu}{\sqrt{2}\sigma}) \right]$. Combining two results, one can easily show that the blue shaded area, corresponding to half of the OVL, or $\mathrm{Err}({x}_0, {y}_0, t)$, is given by:
\begin{equation}
    \mathrm{Err}({x}_0, {y}_0, t) = \frac{1}{2} \mathrm{OVL}\left(q({x}_t | {x}_0), q({y}_t | {y}_0) \right) = \frac{1}{2} \left[ 1 - \mathrm{erf}\left(\frac{ \sqrt{\bar{\alpha}_t} ({y}_0-{x}_0) }{2\sqrt{2(1-\bar{\alpha}_t)}}\right) \right].
\end{equation}
To generalize the results to multi-variate Gaussian distributions, we use the results in~\cite{lu1989multivariate}, which shows that by projecting the data to Fisher's linear discriminant axis, the OVL defined on the discriminant densities is equal to that defined on the multivariate densities. Specifically, the mean of the discriminant densities are given by
\begin{equation}
    {\mu}_0 = \sqrt{\bar{\alpha}_t}(\mathbf{y}_0 - \mathbf{x}_0)^\top \Sigma^{-1} \mathbf{x}_0,\,\, {\mu}_1 = \sqrt{\bar{\alpha}_t}(\mathbf{y}_0 - \mathbf{x}_0)^\top \Sigma^{-1} \mathbf{y}_0,
\end{equation}
where $\Sigma = \beta_t \mathbf{I}$. The common variance of the discriminant densities is given by $\sqrt{\bar{\alpha}_t}(\mathbf{y}_0 - \mathbf{x}_0)^\top \Sigma^{-1} (\mathbf{y}_0 - \mathbf{x}_0)$. Following the calculation steps to compute OVL for the 1-D case, one can show that for both 1-D and multi-variate case, we have
\begin{equation}
    \mathrm{Err}(\mathbf{x}_0, \mathbf{y}_0, t) = \frac{1}{2} \mathrm{OVL}\left(q(\mathbf{x}_t | \mathbf{x}_0), q(\mathbf{y}_t | \mathbf{y}_0) \right) = \frac{1}{2} \left[ 1 - \mathrm{erf}\left(\frac{\lVert \sqrt{\bar{\alpha}_t} (\mathbf{y}_0-\mathbf{x}_0) \rVert}{2\sqrt{2(1-\bar{\alpha}_t)}}\right) \right].
    \label{eq:err}
\end{equation}
As $\bar{\alpha}_t$ decreases with an increasing $t$ from Eq. (2), and the error function $\mathrm{erf}(\cdot)$ is strictly increasing, $\mathrm{Err}(\mathbf{x}_0, \mathbf{y}_0, t)$ is strictly increasing in $t$ given any $\mathbf{x}_0,\mathbf{y}_0$. Hence $\mathop{\mathbb{E}}_{\mathbf{x}_0 \in \mathcal{X}} \left[ \mathrm{Err}(\mathbf{x}_0, \mathbf{y}_0=g_i \cdot \mathbf{x}_0, t) \right] \geq \mathop{\mathbb{E}}_{\mathbf{x}_0 \in \mathcal{X}} \left[ \mathrm{Err}(\mathbf{x}_0, \mathbf{y}_0=g_i \cdot \mathbf{x}_0, t(\mathcal{Z}_i)) \right]$ for every $t\geq t(\mathcal{Z}_i)$, which completes the proof of Theorem 1.

Given that $\mathrm{erf}(\cdot)$ is strictly increasing and $\lVert \mathbf{x}_0 - g_i \cdot \mathbf{x}_0 \rVert$ is first-order stochastic dominant over $\lVert \mathbf{x}_0 - g_j \cdot \mathbf{x}_0 \rVert$, we have $\mathop{\mathbb{E}}_{\mathbf{x}_0 \in \mathcal{X}} \left[ \mathrm{Err}(\mathbf{x}_0, g_i \cdot \mathbf{x}_0, t) \right] > \mathop{\mathbb{E}}_{\mathbf{x}_0 \in \mathcal{X}} \left[ \mathrm{Err}(\mathbf{x}_0, g_j \cdot \mathbf{x}_0, t) \right]$ at every time-step $t$ using Eq.~\ref{eq:err}. Hence $t(\mathcal{Z}_i) > t(\mathcal{Z}_j)$ under any variance schedule $\{\beta_i\}_{i=1}^T$ such that $\mathcal{Z}_i$ is \emph{not} lost at $t(\mathcal{Z}_j)$, completing the proof of Theorem 2.

\textbf{Disentangled Representation Minimizes Eq. (6)}. Suppose that we have a disentangled representation $f$ that maps images to $\{ \mathbf{z}_i \}_{i=1}^T$. Without loss of generality, we assume an attribute order condition where $\mathbf{z}_1,\ldots,\mathbf{z}_T$ take the order such that $\{\mathbf{z}_i\}_{i=1}^t$ makes up the cumulatively lost attributes at each $t$, \ie, $t(\mathcal{Z}_i) \leq t, \forall i \in \{1,\ldots,t\}$ and $t(\mathcal{Z}_i) \geq t, \forall i \in \{t,\ldots,T\}$. Hence given $t$, for each $g_i$ such that $\mathop{\mathbb{E}}_{\mathbf{x}_0 \in \mathcal{X}} \left[ \mathrm{Err}(\mathbf{x}_0, \mathbf{y}_0=g_i \cdot \mathbf{x}_0, t) \right] \geq \tau$, we have $[f(\mathbf{x}_0)]_t \neq [f(g_i \cdot \mathbf{x}_0)]_t, \, \forall \mathbf{x}_0 \in \mathcal{X}$, where $[\cdot]_t$ extracts $\{\mathbf{z}_i\}_{i=1}^t$ from $\{ \mathbf{z}_i \}_{i=1}^T$. Hence there exists an decoder $g$ that maps each unique $[f(\mathbf{x}_0)]_t$ to the corresponding reconstruction error $\hat{\mathbf{x}}_0 - \mathbf{x}_0$ by the pre-trained DM. For other $g_i$, the reconstruction error is bounded by $\mathop{\mathbb{E}}_{\mathbf{x}_0 \in \mathcal{X}} \left[ \mathrm{Err}(\mathbf{x}_0, \mathbf{y}_0=g_i \cdot \mathbf{x}_0, t) \right] < \tau$. 
Hence we prove that the reconstruction error (or attribute loss) can be arbitrarily small (up to specified $\tau$) given a disentangled representation $f$ and a variance schedule that satisfies Theorem 2 (to make sure the attribute order condition holds).
\section{Additional Experiment Details}
\label{sec:D}

\textbf{Network Architecture}. We exactly follow the encoder and decoder design in PDAE~\cite{zhang2022unsupervised} and use the same pre-trained DM. Please refer to PDAE for more details.

\begin{wrapfigure}{r}{0.4\textwidth}
\captionsetup{font=footnotesize,labelfont=footnotesize}
    \centering
    \includegraphics[width=.9\linewidth]{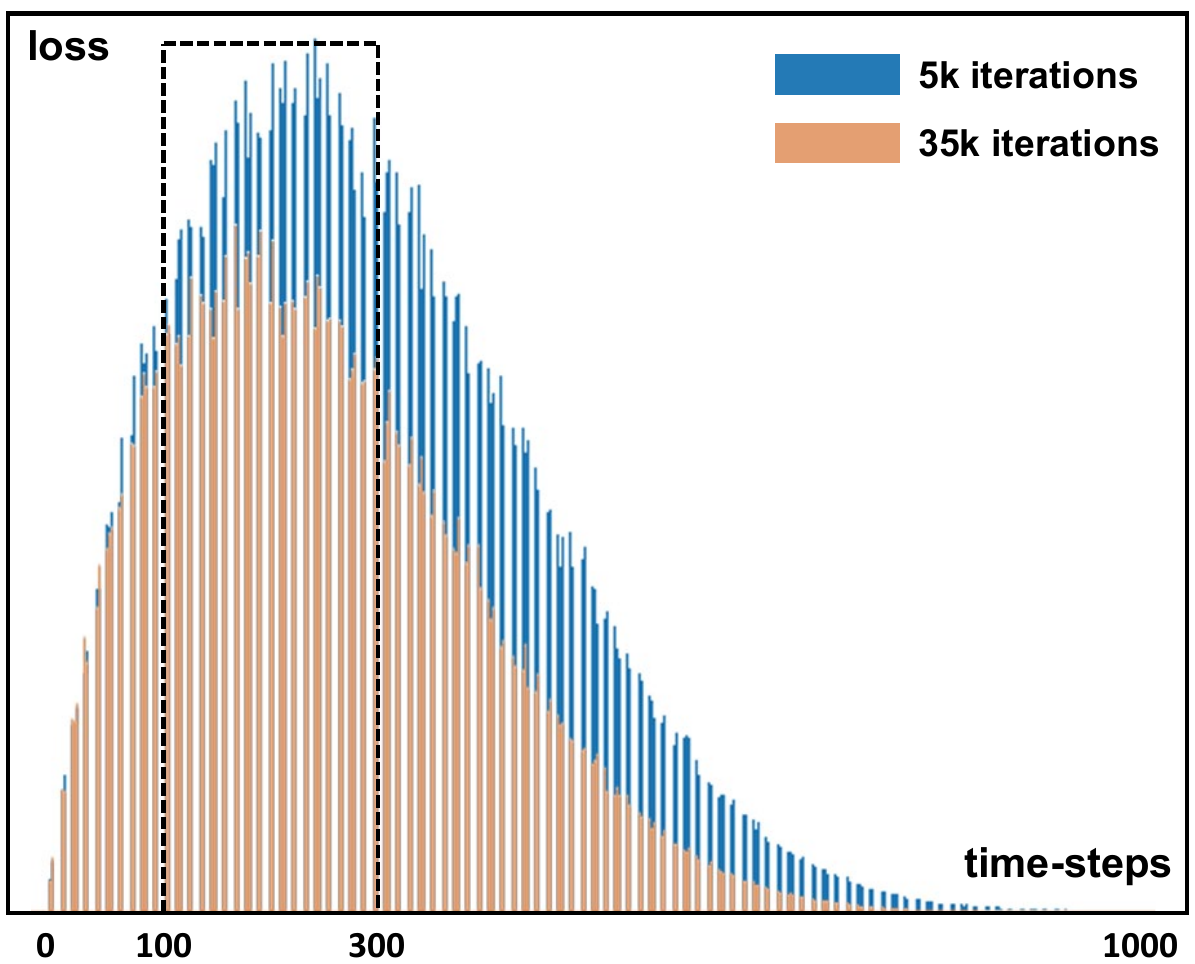}
    
    \caption{Average $\mathcal{L}_t$ for each time-step $t$ in DiTi training at 5k iterations and 35k iterations.}
    \label{fig:loss_at_t}
    \vspace{-2mm}
\end{wrapfigure}

\textbf{Imbalanced Partition Strategy}. As shown in Figure~\ref{fig:loss_at_t}, we plot the average loss $\mathcal{L}_t$ (in the most recent 5k iterations) at each time-step $t$. It is clear that time-step 100-300 contribute the most to the overall loss. Furthermore, by comparing the loss at 5k iteration and 35k iteration, we observe that the same time-step range contributes the most to the loss minimization. We conjecture that the time-step 100-300 contains rich semantic information.
On the other hand, late time-steps (\eg, after $t=500$) have smaller loss value and less loss reduction, as late time-steps have very small weight $\lambda_t$ by the design of DDPM~\cite{ho2020denoising}.
Hence accordingly, we design an imbalanced
partition strategy to allocate more feature dimensions to time-step 100-300 and less ones to time-step 500-1000.
Specifically, we assign 10, 25, 327, 100, 50 dimensions to time-step range 0-50, 50-100, 100-300, 300-500, 500-1000, respectively.
Note that we only tried this dimension allocation strategy as a heuristic approach, and we did not search for an optimal strategy.
Future work can explore an adaptive allocation strategy.

\begin{wrapfigure}{r}{0.4\textwidth}
  \centering
  \includegraphics[width=\linewidth]{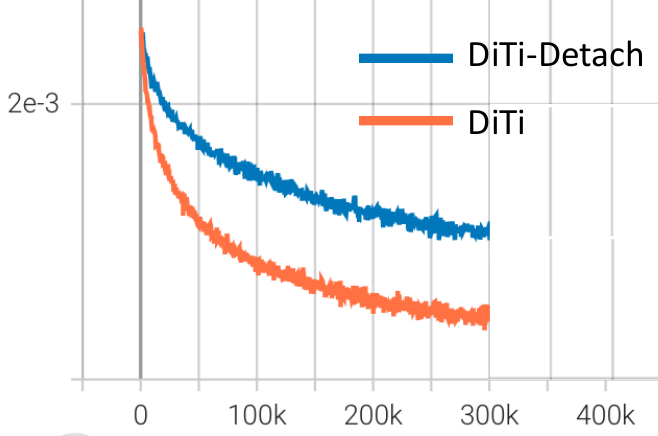}
  \caption{Training loss of DiTi and DiTi with detach optimiaztion strategy.}
  \label{suppfig:loss-detach}
\end{wrapfigure}

\textbf{Optimization Strategy}. Figure~\ref{suppfig:loss-detach} compares the loss in Eq. (6) by DiTi and DiTi-Detach (\ie, detaching the gradients of $\mathbf{z}_1,\ldots,\mathbf{z}_{t-1}$) throughout training. The loss reduction of DiTi-Detach is much slower as only $\frac{1}{k}$ of the feature is trained at each iteration.
As shown in Table 2, this alternative optimization strategy hurts the performance when transferring the feature trained on
CelebA for LFW attribute regression. We conjecture that transfer learning and regression task is more difficult, hence LFW regression is more sensitive to model convergence.
However, this strategy does provide additional inductive bias towards disentanglement, as only $\mathbf{z}_t$ is trained to capture the lost attribute at time-step $t$.
Hence we use DiTi for classification/regression tasks and DiTi-Detach for generation tasks.
As future work, we will explore improved network design and other optimization techniques to reap the benefits of DiTi-Detach strategy without hurting convergence.

\textbf{Training Algorithm}. Please refer to Algorithm~\ref{alg:train}.

\textbf{Counterfactual Generation Algorithm}. Please refer to Algorithm~\ref{alg:cf}.

\textbf{Extended Manipulation Algorithm}. We use the attribute labels to train a linear classifier that predicts a specific attribute. On CelebA~\cite{liu2015faceattributes}, we use its 40 attribute labels for training. On Bedroom~\cite{yu2015lsun}, there are no ground-truth attribute labels. We use pseudo-labels produced by an off-the-shelf attribute predictor~\cite{zhou2017places} to train the attribute classifier.
In particular, we adopt ProbMask~\cite{zhou2021effective} to constrain the classifier such that its weight has only $d'$ non-zero dimensions, where $d' < d$ (\eg, $d'=$ 16 or 32 and $d=512$). This design is to test the modularity of the feature---a specific attribute (\eg, ``Young'') should be captured by the combination of a few modular attributes $\mathbf{z}_i$, but not all.
With the trained classifier for an attribute, to manipulate the attribute with scale $\lambda$ on a sample $\mathbf{x}_0$, we first obtain its feature $\mathbf{z}=f(\mathbf{x}_0)$, then push $\mathbf{z}$ along the normal vector of the decision boundary with certain scale $\lambda$, resulting in manipulated code $\mathbf{z}'$, and finally encode $\mathbf{x}_T$ back to the manipulated image by the guidance of $g(\mathbf{z}',t)$.
The process is summarized in Algorithm~\ref{alg:manip}.

\begin{algorithm}[H]
    \SetAlgoLined
    \SetKwInOut{KwIn}{Input}
    \SetKwInOut{KwOut}{Output}
    \KwIn{Training data distribution $q(\mathbf{x}_0)$, pre-trained DM $u_\theta$}
    \KwOut{Trained encoder $f$, decoder $g$}
     Randomly initialize $f,g$\;
     \While{not converged}{
        $\mathbf{x}_0\sim q(\mathbf{x}_0)$\;
        $\mathbf{z}=f(\mathbf{x}_0)$\;
        Partition $\mathbf{z}$ into $\{\mathbf{z}_i\}_{i=1}^T$\;
        $t\sim \mathrm{Uniform}(1,\ldots,T)$\;
        $\boldsymbol{\epsilon} \sim \mathcal{N}(\mathbf{0},\mathbf{I})$\;
        $\mathbf{x}_t = \sqrt{\bar{\alpha}_t} \mathbf{x}_0 + \sqrt{1- \bar{\alpha}_t} \boldsymbol{\epsilon}$\;
        $\bar{\mathbf{z}}_t = [\mathbf{z}_1,\ldots,\mathbf{z}_t,\mathbf{0},\ldots,\mathbf{0}]$\;
        Update $f,g$ by minimizing $\lambda_t \lVert \mathbf{x}_0 - \left(u_\theta(\mathbf{x}_t, t) + w_t g(\bar{\mathbf{z}}_t, t) \right) \rVert^2$ in Eq. (6)\;
     }
     \KwRet{$f,g$}
     \caption{DiTi training}
     \label{alg:train}
\end{algorithm}

\begin{algorithm}[H]
    \SetAlgoLined
    \SetKwInOut{KwIn}{Input}
    \SetKwInOut{KwOut}{Output}
    \KwIn{$\mathbf{x}_0$, $\mathbf{x}'_0$, subset $\mathcal{S}\subset \{1,\ldots,k\}$, scale $\lambda$, pre-trained $u_\theta$, trained $f,g$, sampling sequence $\{t_i\}_{i=1}^M$ where $t_1=0$ and $t_M=T$}
    \KwOut{A counterfactual image for $\mathbf{x}_0$}
     Compute $\mathbf{x}_T,\mathbf{x}'_T$ for $\mathbf{x}_0,\mathbf{x}'_0$ with DDIM inversion, respectively\;
     $\mathbf{z}=f(\mathbf{x}_0), \mathbf{z}'=f(\mathbf{x}'_0)$\;
     Partition $\mathbf{z}$ into $\{\mathbf{z}_i\}_{i=1}^T$, $\mathbf{z}'$ into $\{\mathbf{z}'_i\}_{i=1}^T$\;
     $\mathbf{z}_\mathcal{S} \leftarrow \mathrm{lerp}(\mathbf{z},\mathbf{z}',\mathcal{S};\lambda)$, \ie, perform linear interpolation on all $\mathbf{z}_i, i\in \mathcal{S}$\; 
     $\mathbf{x}_T \leftarrow \mathrm{slerp}(\mathbf{x}_T,\mathbf{x}'_T;\lambda)$\;
     \For{$i=M,\ldots,2$} {
        $\hat{\mathbf{x}}_0 = u_\theta (\mathbf{x}_{t_i}, t_i) + w_t g(\mathbf{z}_\mathcal{S}, t_i)$\;
        $\mathbf{x}_{t_{i-1}} \leftarrow \sqrt{\bar{\alpha}_{t_{i-1}}} \hat{\mathbf{x}}_0 +  \frac{\sqrt{1- \bar{\alpha}_{t_{i-1}}} (\mathbf{x}_{t_i} - \sqrt{\bar{\alpha}_{t_i}} \hat{\mathbf{x}}_0) }{\sqrt{1- \bar{\alpha}_{t_i}}}$\;
     }
     \KwRet{$\mathbf{x}_0$}
     \caption{Counterfactual generation from $\mathbf{x}_0$ to $\mathbf{x}'_0$ on subset $\mathcal{S}$ with scale $\lambda$}
     \label{alg:cf}
\end{algorithm}


\begin{algorithm}[H]
    \SetAlgoLined
    \SetKwInOut{KwIn}{Input}
    \SetKwInOut{KwOut}{Output}
    \KwIn{Original $\mathbf{x}_0$, manipulation scale $\lambda$, trained ProbMask classifier with weight parameter $\mathbf{w}\in \mathbb{R}^d$, pre-trained DM $u_\theta$, trained $f,g$, standard deviation $\boldsymbol{\sigma}$ of $\mathbf{z}$ in the entire training dataset, sampling sequence $\{t_i\}_{i=1}^M$ where $t_1=0$ and $t_M=T$}
    \KwOut{A counterfactual image for $\mathbf{x}_0$}
     Compute $\mathbf{x}_T$ for $\mathbf{x}_0$ with DDIM inversion\;
     $\mathbf{z}=f(\mathbf{x}_0)$\;
     $\mathbf{z}' = \mathbf{z} + \lambda \frac{\boldsymbol{\sigma} \cdot \mathbf{w}}{\lVert \mathbf{w} \rVert}$\;
     \For{$i=M,\ldots,2$} {
        $\hat{\mathbf{x}}_0 = u_\theta (\mathbf{x}_{t_i}, t_i) + w_t g(\mathbf{z}', t_i)$\;
        $\mathbf{x}_{t_{i-1}} \leftarrow \sqrt{\bar{\alpha}_{t_{i-1}}} \hat{\mathbf{x}}_0 +  \frac{\sqrt{1- \bar{\alpha}_{t_{i-1}}} (\mathbf{x}_{t_i} - \sqrt{\bar{\alpha}_{t_i}} \hat{\mathbf{x}}_0) }{\sqrt{1- \bar{\alpha}_{t_i}}}$\;
     }
     \KwRet{$\mathbf{x}_0$}
     \caption{Manipulating $\mathbf{x}_0$ with a trained ProbMask classifier and scale $\lambda$}
     \label{alg:manip}
\end{algorithm}

\section{Additional Experiment Results}
\label{sec:E}

\begin{figure}[t!]
    \centering
    \captionsetup{font=footnotesize,labelfont=footnotesize,skip=5pt}
    \includegraphics[width=1.0\linewidth]{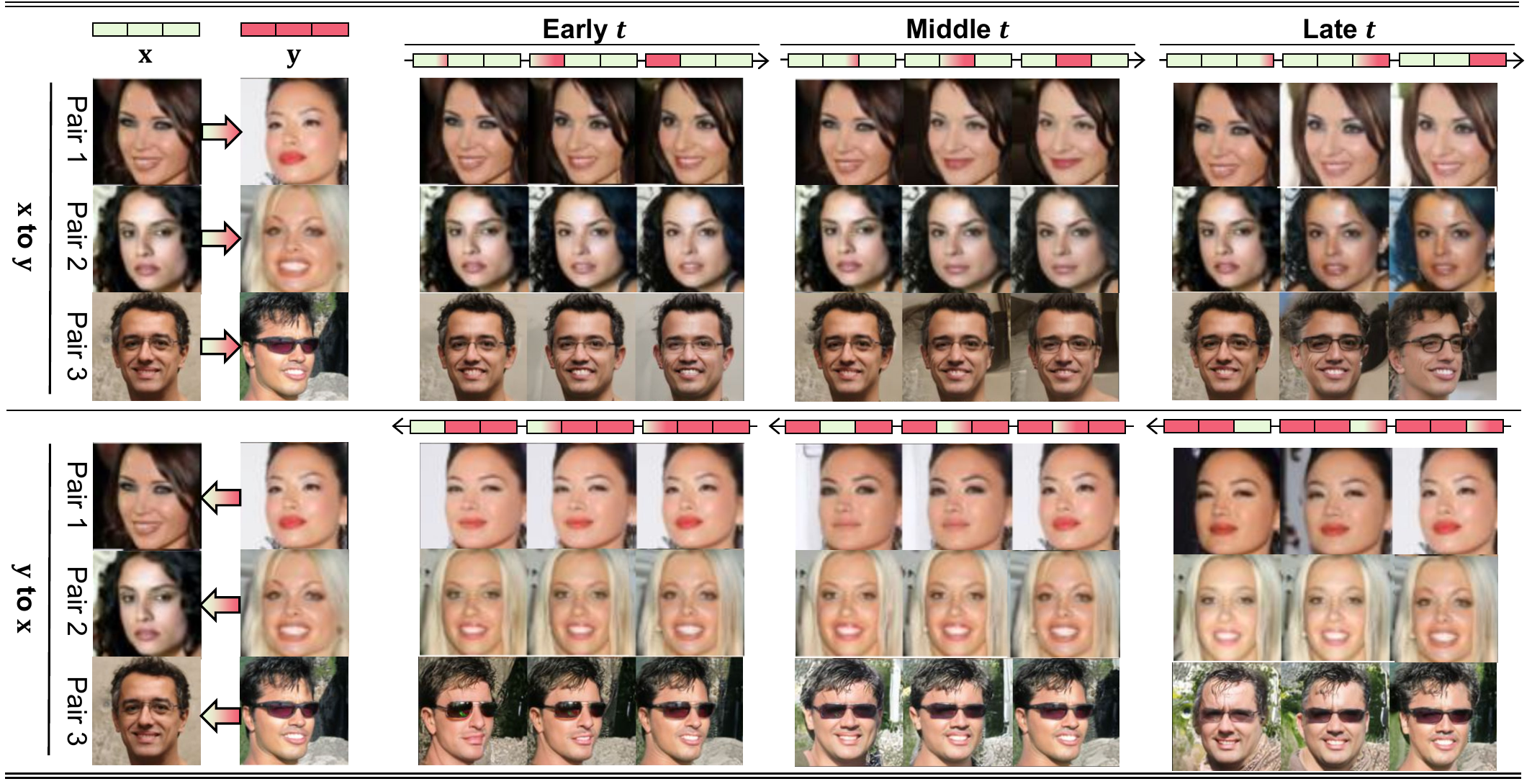}
    \caption{Supplementary to Figure~\ref{fig:5}. More counterfactual generations by our DiTi.}
    \label{fig:a5_diti}
    \vspace*{-4mm}
\end{figure}

\begin{figure}[t!]
    \centering
    \captionsetup{font=footnotesize,labelfont=footnotesize,skip=5pt}
    \includegraphics[width=1.0\linewidth]{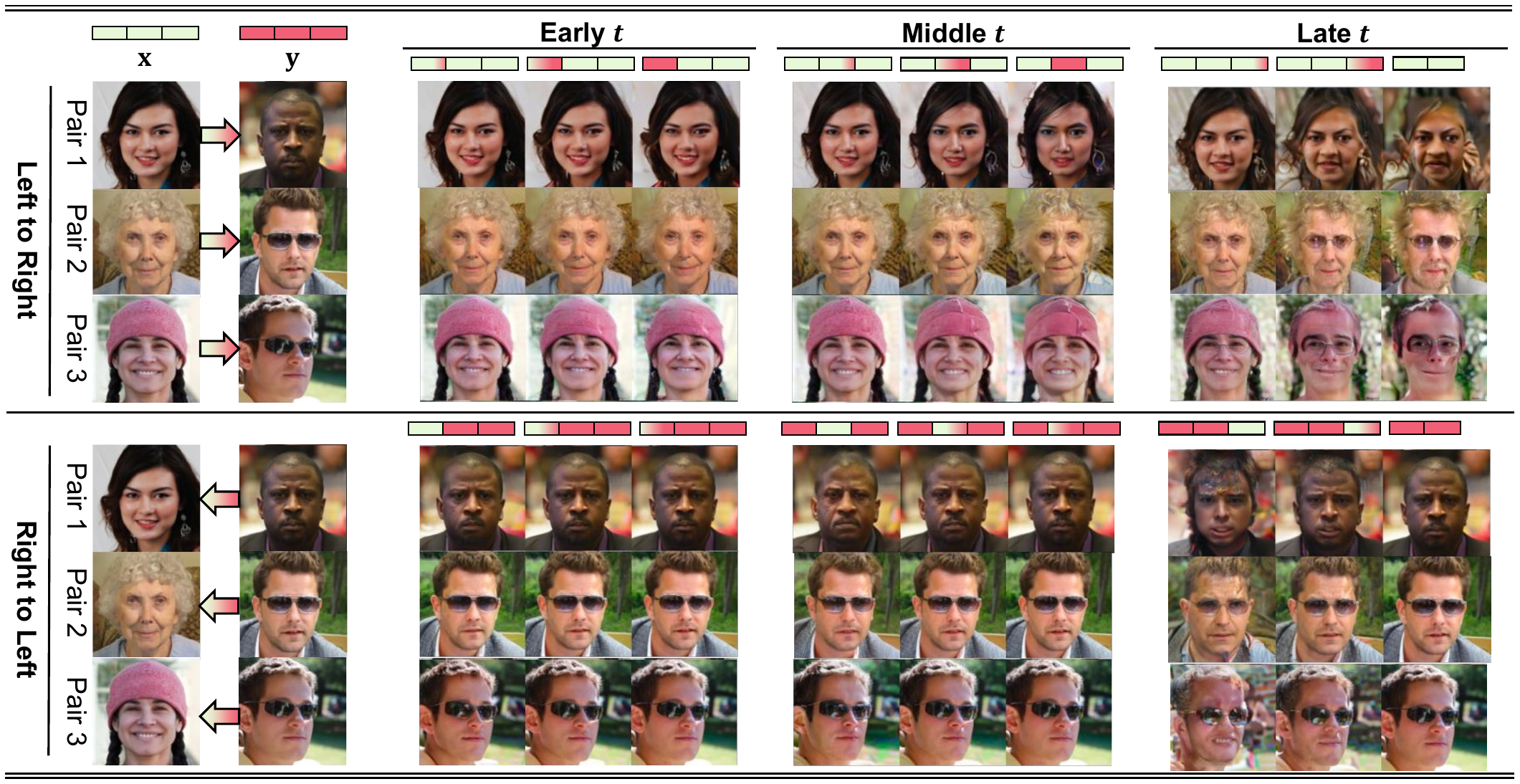}
    \caption{Supplementary to Figure~\ref{fig:5} by using Diff-AE.}
    \label{fig:a5_diffae}
    \vspace*{-4mm}
\end{figure}

\begin{figure}[t!]
    \centering
    \captionsetup{font=footnotesize,labelfont=footnotesize,skip=5pt}
    \includegraphics[width=1.0\linewidth]{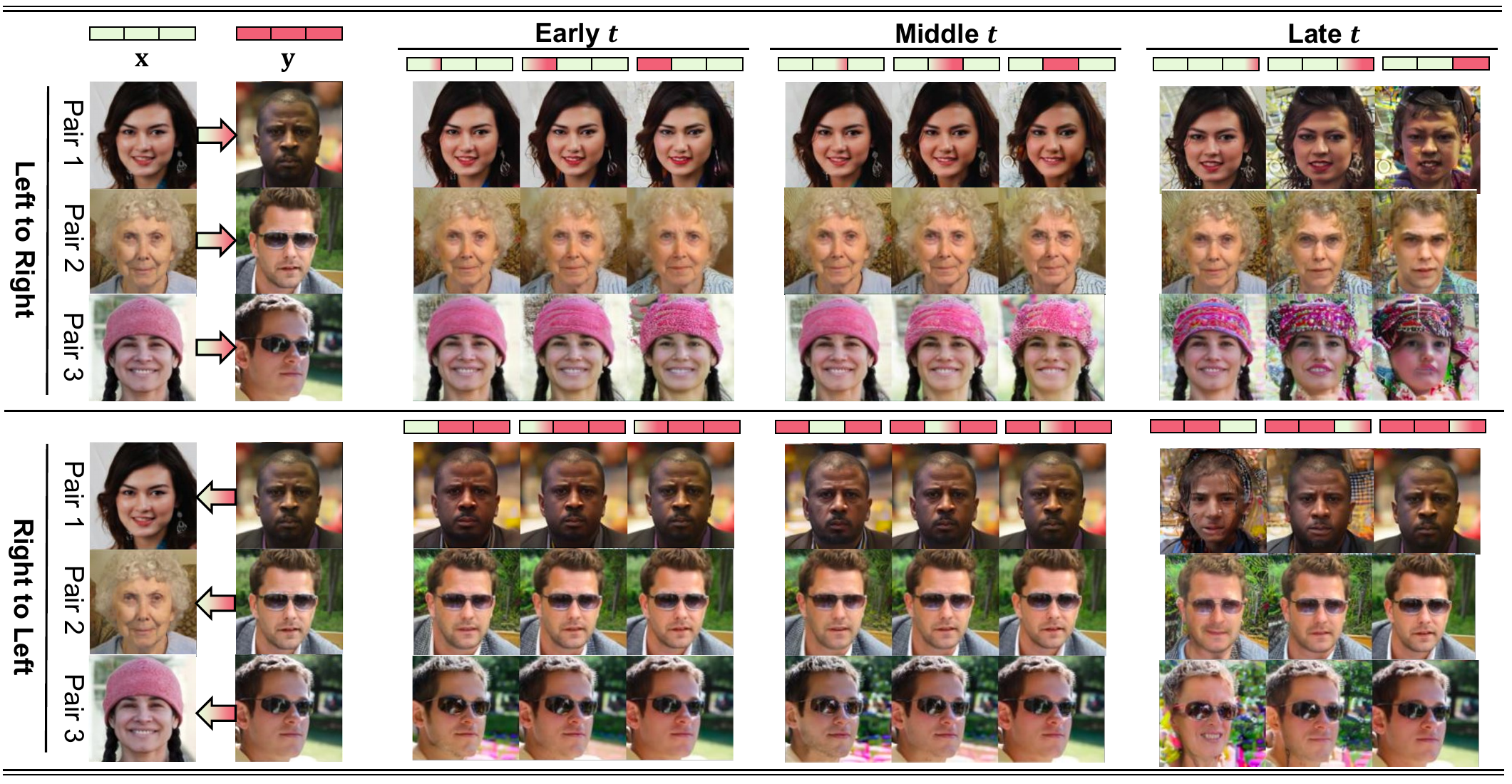}
    \caption{Supplementary to Figure~\ref{fig:5} by using PDAE.}
    \label{fig:a5_pdae}
    \vspace*{-4mm}
\end{figure}

\textbf{SimCLR Results without Augmentation}. By removing color-related augmentations (\ie, color jittering and random grayscale), SimCLR (trained on CelebA) suffers severe performance degradation, only obtaining 34.7\% AP, 0.176 Pearson's r and 0.717 MSE.

\textbf{More Counterfactual Generations by Interpolation}. In Figure~\ref{fig:a5_diti}, we show additional counterfactual generations by DiTi.

\textbf{Baseline Counterfactual Generations by Interpolation}. In Figure~\ref{fig:a5_diffae} and Figure~\ref{fig:a5_pdae}, we show the baseline results under the same setting as Figure~\ref{fig:5}. In contrast to our DiTi, baselines either make no meaningful edit or produce distorted counterfactual generations, which suggest that their features are still entangled.

\begin{figure}[t!]
    \centering
    \captionsetup{font=footnotesize,labelfont=footnotesize,skip=5pt}
    \includegraphics[width=1.0\linewidth]{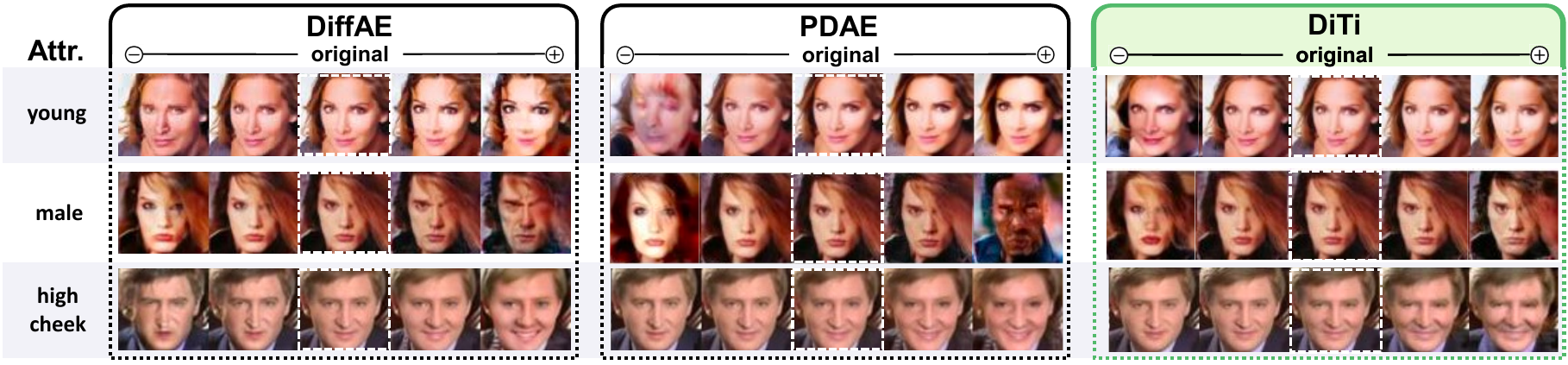}
    \caption{Counterfactual generations on CelebA by manipulating 32 out of 512 feature dimensions. Supplementary to Figure~\ref{fig:manipulate}.}
    \label{fig:a2_1}
    \vspace*{-4mm}
\end{figure}
\begin{figure}[t!]
    \centering
    \captionsetup{font=footnotesize,labelfont=footnotesize,skip=5pt}
    \includegraphics[width=1.0\linewidth]{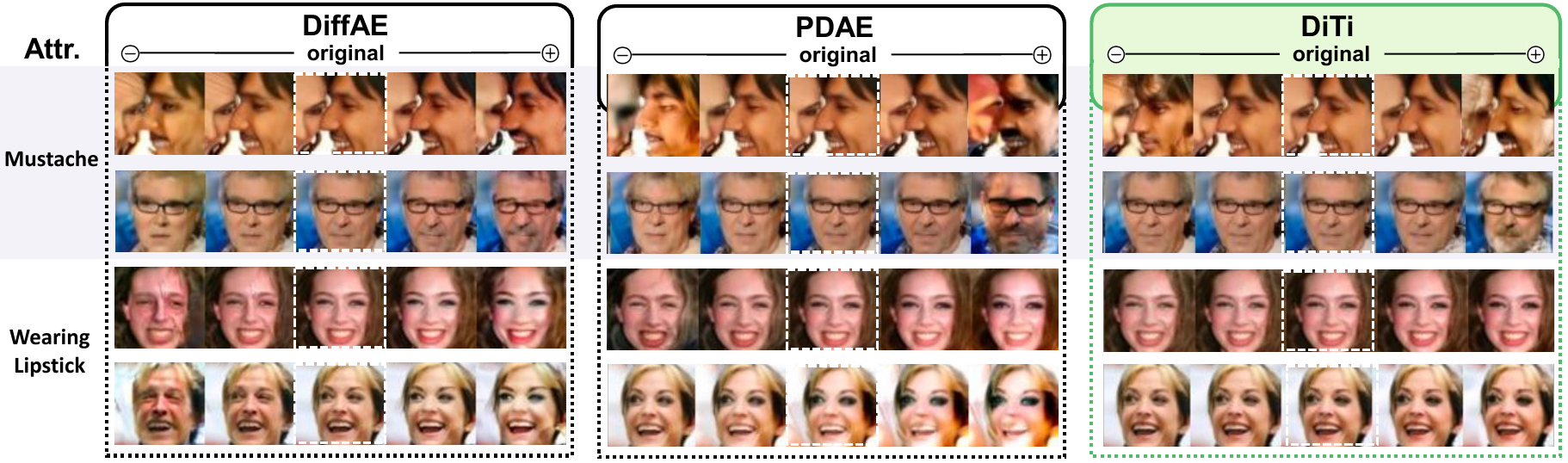}
    \caption{CelebA counterfactual generations on two other attributes by manipulating 32 out of 512 feature dimensions. Supplementary to Figure~\ref{fig:manipulate}.}
    \label{fig:a2_2}
    \vspace*{-4mm}
\end{figure}
\begin{figure}[t!]
    \centering
    \captionsetup{font=footnotesize,labelfont=footnotesize,skip=5pt}
    \includegraphics[width=1.0\linewidth]{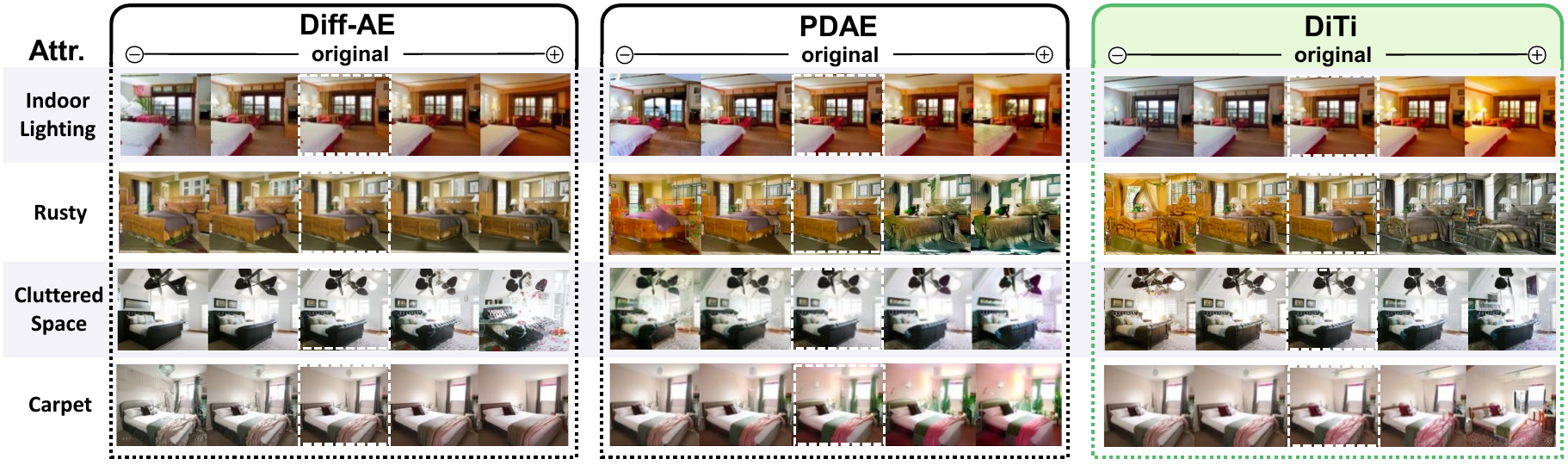}
    \caption{More counterfactual generations on Bedroom by manipulating 32 out of 512 feature dimensions. Supplementary to Figure~\ref{fig:6}.}
    \label{fig:a6}
    \vspace*{-4mm}
\end{figure}

\begin{figure}[t!]
    \centering
    \captionsetup{font=footnotesize,labelfont=footnotesize,skip=5pt}
    \includegraphics[width=1.0\linewidth]{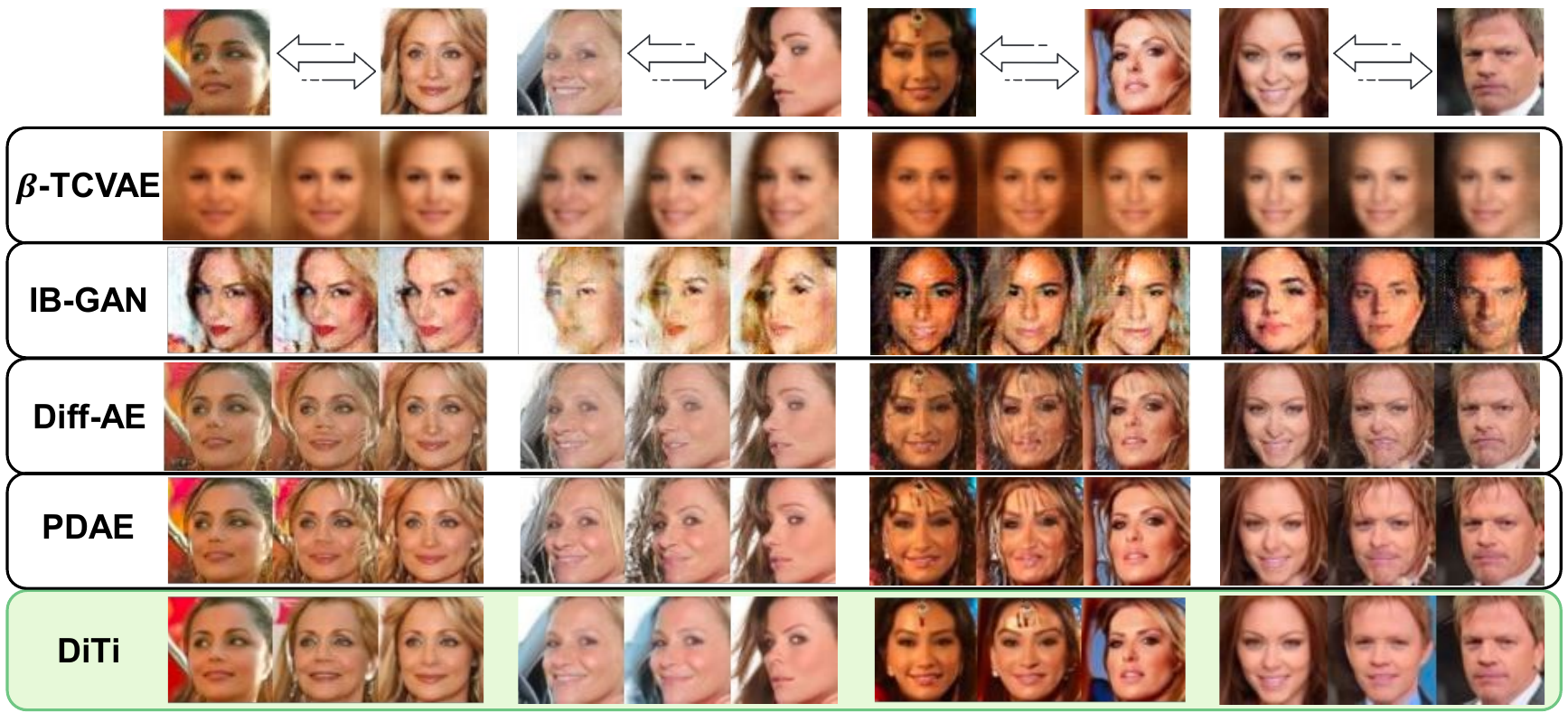}
    \caption{Results of interpolating the whole feature on CelebA. Baselines have distortions during interpolation. Supplementary to Figure~\ref{fig:7}.}
    \label{fig:a7}
    \vspace*{-4mm}
\end{figure}

\textbf{More CelebA Manipulations}. In Figure~\ref{fig:a2_1}, we compare the results of Diff-AE with PDAE and DiTi on CelebA manipulation. In Figure~\ref{fig:a2_2}, we include more CelebA manipulation results.

\textbf{More Bedroom Manipulations}. In Figure~\ref{fig:a6}, we show more results of Diff-AE, PDAE and DiTi on Bedroom manipulation.

\textbf{More Whole-feature Interpolations}. In Figure~\ref{fig:a7}, we show interpolation results on CelebA of all generative baselines and our DiTi. Only DiTi can interpolate without artifacts or distortions. Notably, $\beta$-TCVAE and IB-GAN have poor generation fidelity and struggle to preserve the face identity. This is in line with the common view of their limitations, \eg, VAE tends to generate blurry images and GAN suffers from unstable training and difficulties of convergence. We postulate that their information bottleneck also limits the faithfulness of their feature.

\end{document}